\documentclass[journal]{IEEEtran}
\usepackage{cite}

\ifCLASSINFOpdf
\else
\fi
\usepackage{graphicx}
\usepackage{float}
\usepackage{subfigure}
\usepackage{amsmath}
\usepackage{algpseudocode}
\usepackage{stfloats}
\usepackage{multirow} 
\usepackage{xcolor}
\usepackage[ruled,vlined]{algorithm2e}
\usepackage{amssymb}
\usepackage{booktabs}
\usepackage{cite}
\makeatletter
\def\BState{\State\hskip-\ALG@thistlm}
\makeatother

\begin{document}

%
\title{FG-Net: Fast Large-Scale LiDAR Point Clouds Understanding Network Leveraging Correlated Feature Mining and Geometric-Aware Modelling}
%
%
%

\author{Kangcheng Liu$^{1}$,~\IEEEmembership{Student Member,~IEEE}, Zhi Gao$^{2}$, Feng Lin$^{3}$, and Ben M. Chen$^{1}$,~\IEEEmembership{Fellow,~IEEE}
        
\thanks{Manuscript received December 16, 2020.  \textit{(Corresponding authors: Kangcheng Liu, Zhi Gao, and Feng Lin)}. 

$^{1}$K. Liu and B. M. Chen are with the Department of Mechanical and Automation Engineering, The Chinese University of Hong Kong, Shatin, N.T.,  Hong Kong 999077, China.
(email: kcliu@mae.cuhk.edu.hk, bmchen@cuhk.edu.hk)}

\thanks{$^{2}$Z. Gao is with the School of Remote Sensing and Information Engineering, Wuhan University, Hubei 430070, China.
(email: gaozhinus@gmail.com)}

\thanks{$^{3}$F. Ling is with the Peng Cheng Laboratory, First Xingke Street, Nanshan, Shenzhen, Guangdong 518066, China.
(email: truelinfeng@outlook.com)}
}



%
%

\markboth{Journal Version, MM~YY}%
{Shell \MakeLowercase{\textit{et al.}}: Bare Demo of IEEEtran.cls for IEEE Journals}
%



\maketitle


\begin{abstract}
This work presents FG-Net, a general deep learning framework for large-scale point clouds understanding without voxelizations, which achieves accurate and real-time performance with a single NVIDIA GTX 1080 GPU. First, a novel noise and outlier filtering method is designed to facilitate subsequent high-level tasks. For effective understanding purpose, we propose a deep convolutional neural network leveraging correlated feature mining and deformable convolution based geometric-aware modelling, in which the local feature relationships and geometric patterns can be fully exploited. For the efficiency issue, we put forward an inverse density sampling operation and a feature pyramid based residual learning strategy to save the computational cost and memory consumption respectively. Extensive experiments on real-world challenging datasets demonstrated that our approaches outperform state-of-the-art approaches in terms of accuracy and efficiency. Moreover, weakly supervised transfer learning is also conducted to demonstrate the generalization capacity of our method. 
\end{abstract}


%
\IEEEpeerreviewmaketitle

\section{Introduction}
%
%
%
%
\IEEEPARstart 
{D}{ue} to the directness and robustness in obtaining 3D information, there has been an increasing proliferation of light detection and ranging (LiDAR) sensors which have been popularly deployed on a variety of intelligent agents such as unmanned ground vehicles (UGVs), unmanned aerial vehicles (UAVs) to perform localization, obstacle detection, exploration, etc. Consequently, efficient and effective large-scale 3D LiDAR point clouds understanding is of great importance to facilitate machine perception, which bridges the gap between 3D points and any high-level information, structural or semantic, or both \cite{wang2019re,cong2018speedup, zhang2019shellnet,guo20143d}. However, due to the electrical and mechanical disturbances, and the reflectance property of targets, the point clouds often suffer from noise and outliers. Moreover, compared with the 2D raster image,
\begin{figure}[htbp!]
\centering
\includegraphics[scale=0.26]{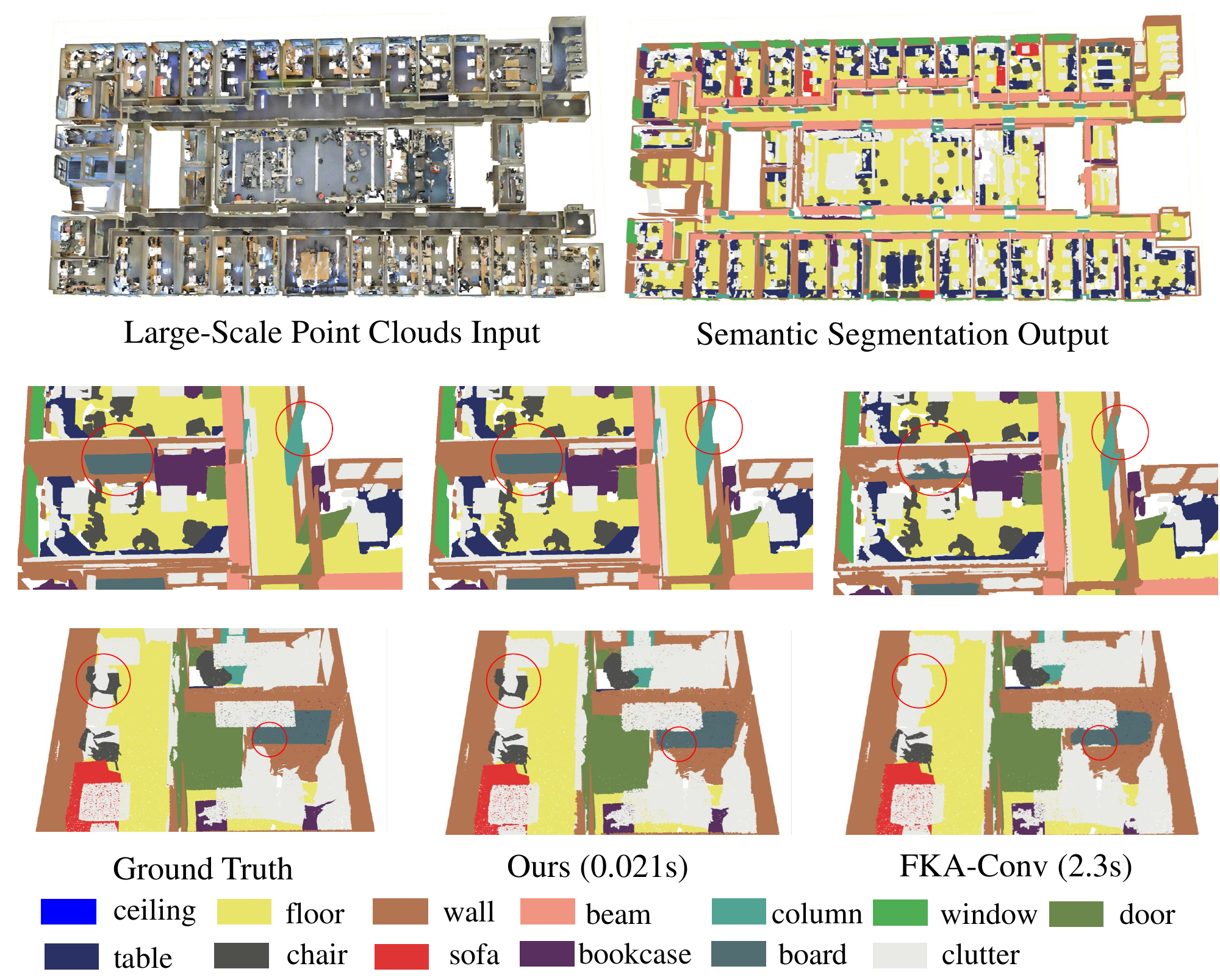}
\caption{Semantic segmentation results of our method compared with the state-of-the-art method FKA-Conv\cite{boulch2020fka} on S3DIS\cite{armeni20163d}. The top row shows the overall segmentation performance by our method. The bottom 2 rows show the detailed comparisons of segmentation performance highlighted by red circles. Our method achieves real-time segmentation performance of 0.021s per $10^5$ points, which is better and faster than the state-of-the-art method FKA-Conv.}
\label{fig_demo}
\vspace{-0.38cm}
\end{figure}
the topological relationship between objects in 3D point clouds is much weakened, rendering the task of segmentation and understanding much more challenging. Therefore, autonomous large-scale point clouds understanding remains an open problem and requires urgent efforts to tackle the challenges, especially when both accuracy and efficiency are taken into account.

Like any high-level task in 2D image domain such as object detection, segmentation, or classification, the point clouds understanding methods can also be classified into traditional category and deep learning based methods. In the traditional category, the representative histogram-based methods \cite{rusu2008aligning} \cite{rusu2009fast} encode the k-Nearest-Neighbor (kNN) geometric features of a 3D point via calculating its surrounding multidimensional average curvature for local geometric variations descriptions. In signature-based \cite{tombari2010unique} and transform-based \cite{guo20143d} methods, handcrafted feature descriptions of point clouds have been proposed and exploited for semantic understanding. However, the performances of these methods are merely demonstrated in well-controlled conditions with ideal assumptions such as noise-free and homogeneous environments\cite{guo20143d}. On the other hand, 
 deep learning based point clouds processing methods have been proposed with promising results in recent years. 
The mainstream point clouds understanding methods can be roughly divided into three categories: projection-based \cite{milioto2019rangenet++,wu2019squeezesegv2,xu2020squeezesegv3,feng2018gvcnn,kundu2020virtual,li2020end,gojcic2020learning}, voxel-based\cite{graham20183d,choy20194d,tang2020searching, meng2019vv,ye2020hvnet}, and direct point-based \cite{landrieu2019point,wang2019re,nezhadarya2020adaptive,zhang2019shellnet, qi2017pointnet++,yan2020pointasnl,liu2019densepoint,qi2017pointnet,shen2018mining,thomas2019kpconv,lei2020spherical, liu2019point, wu2019pointconv}. Such representative works of each category will be discussed in Section II, and the limitations of these methods are summarized here: First, the sampling operation of almost all these methods has high computational cost and memory consumption. For instance, the widely applied farthest point sampling (FPS) \cite{qi2017pointnet++} \cite{yan2020pointasnl} takes more than 1000 seconds to subsample $10^5$ points to about $10^3$ points. Furthermore, their subsequent perception networks usually rely on the expensive operations, such as voxelizations \cite{graham20183d} \cite{choy20194d}, or graph construction \cite{landrieu2019point}, etc. Second, nearly all the existing methods are designed for small-scale point clouds without considering noise and outliers which are inevitable in practice. Moreover, the large-scale point clouds typically suffer from great class imbalance in semantic categories, and the points obtained by LiDAR in complex dynamic environments are often irregular, orderless, and have distant distributed semantic information. For example, in autonomous driving scenarios, the typical objects exhibit diverse geometric shapes with varying object sizes (e.g., cyclists and persons) or have distribution across a long spatial range in a non-uniform way (e.g., road, buildings, and vegetations). However, to the best of our knowledge, the existing methods can hardly capture the complex geometry and the latent feature correlations in large-scale point clouds effectively.

To overcome the aforementioned challenges, we propose a general deep learning framework named \textit{\textbf{FG-Net}} for large-scale point clouds understanding. We adopt deformable convolution for modelling the geometric structure, and pointwise attentional aggregation for mining the correlated features among point clouds. It should be noted that the deformable convolutional modelling can effectively adapt to the local geometry of objects by deformed kernels that dynamically adapt to diverse local geometries while the correlated feature mining can capture the distributed contextual information in spatial locations and semantic features adaptively across a long spatial range. The modules in our network can be implemented with simple pointwise matrix multiplication and add operations, which can be easily parallelized by GPU for acceleration. As shown in Fig. \ref{fig_demo}, our method outperforms state-of-the-art ones in terms of both accuracy and efficiency, rendering it achievable to realize real-time perception performance on the large-scale point clouds. In summary, our work makes the following contributions:
\begin{enumerate}[]
\setlength{\itemsep}{0pt}
\setlength{\parskip}{0pt}
\setlength{\parsep}{0pt}
\item We propose pointwise correlated feature mining and geometric-aware modelling module for large-scale point clouds understanding.  Furthermore, we interpret the effectiveness of our network by visualizing the complementary features captured by our network modules. 
\item 
  We propose a feature pyramid based residual learning architecture to leverage patterns at different resolutions in a memory-efficient way.  Extensive experiments on real-world challenging datasets demonstrated that our approaches outperform state-of-the-art ones in terms of accuracy and efficiency.
\item 
  We propose a novel fast noise and outliers removal method and a points down-sampling strategy for large-scale point clouds, which simultaneously enhances the performance and improve the efficiency of semantic understanding tasks in the large-scale scene. 
  

\end{enumerate} 
The paper is organized as follows: Section II gives a detailed review of existing methods of point clouds understanding. In Section III, we illustrate our proposed framework thoroughly and gives our optimization function formulation and training details. We also propose point clouds filtering methods and sampling methods which are specially designed to speed up our framework on large-scale point clouds. Section IV gives substantial results of our experiments and ablation studies. Section V concludes our work.


\section{Related Works}
Advanced deep learning techniques in 2D image domain have been investigated extensively, and resulted in stunning performance \cite{yang2020hierarchical, yan2020new, kim2019keyboard,chen2020deep,liu2019weakly}. Naturally, such deep learning techniques have been exploited for point clouds processing and understanding, and the published works can be roughly categorized into voxel-based
\cite{graham20183d,choy20194d,tang2020searching, meng2019vv,ye2020hvnet}, projection-based\cite{milioto2019rangenet++,wu2019squeezesegv2,xu2020squeezesegv3,feng2018gvcnn,kundu2020virtual,li2020end,gojcic2020learning} and point-based methods\cite{landrieu2019point,wang2019re,nezhadarya2020adaptive,zhang2019shellnet, qi2017pointnet++,yan2020pointasnl,liu2019densepoint,qi2017pointnet,shen2018mining,thomas2019kpconv,lei2020spherical, liu2019point, wu2019pointconv}. The voxel-based and projection-based methods transform point clouds into different representations while the point-based methods process point clouds directly. These methods are mainly designed and tested on the relative small-scale point clouds of less than $10^5$ points with block partitioning. Directly extending them to deal with large-scale point clouds will result in prohibitively expensive computational costs. Here we discuss these methods thoroughly of their advantages and shortcoming, and the rationale that motivates our modifications and improvements.


\subsection{Voxel-based and Projection-based Methods for Point Clouds Understanding}

The most recent and typical voxel-based methods are SparseConv \cite{graham20183d} and Minkowski CNN \cite{choy20194d}. The voxel-based methods use 3D convolutions which are intuitive extensions of 2D counterparts. The advantage of them is that the spatial relationship can be well reserved with high voxel resolutions, but these methods are quite computationally expensive. The computation cost and memory consumption of the voxel-based models increase cubically with the resolution of input point clouds. By contrast, the geometric information loss will be 
significant if we decrease the resolution of voxelization. In addition, the voxel-based methods rely on aggressive down sampling to achieve bigger receptive fields, resulting in even lower resolution in deeper network layers. Hence, it is quite hard to achieve real-time performance while considering the balance between accuracy and computational cost. The projection is also used to project point clouds into range images\cite{milioto2019rangenet++,wu2019squeezesegv2,xu2020squeezesegv3} or multi-view images\cite{feng2018gvcnn,kundu2020virtual,li2020end,gojcic2020learning}, to facilitate the use of 2D CNNs. However, such projection inevitably leads to the loss of geometrical information. In practice of dealing with large-scale point clouds, the drawbacks of voxel-based and projection-based methods become more prohibitive.

\subsection{Point-based Methods for Point Clouds Understanding} 
The pointNet \cite{qi2017pointnet} is the pioneering work that extracts the pointwise feature directly using shared Multi-layer Perceptron (MLP). It is permutation invariant to point clouds orders because it uses max-pooling operations. The pointNet++\cite{qi2017pointnet++} extracts the local features using pointNet and considers local geometric relationships with the hierarchical grouping and abstraction as well as the multi-scale and resolution grouping. More point-based methods\cite{li2018pointcnn,nezhadarya2020adaptive,landrieu2019point} have been proposed recently with complicated network design to aggregate local features. 
However, all these methods are not able to model intrinsic geometric structures of points or to capture the non-local distributed contextual correlations in spatial locations and semantic features effectively. There are also a series of new explorations on how to implement convolution on point clouds. The methods \cite{shen2018mining,thomas2019kpconv,lei2020spherical} focus on how to learn kernels which can better capture the local geometry of points. However, the proposed convolutional kernels are too complicated to be directly applied to deep neural networks for large-scale point clouds understanding. Motivated by the challenges above, we proposed a novel lightweight point-based method to consider distributed long-range dependencies and learn kernels to capture the local structures of point clouds. 
\subsection{Efficient Large-Scale Point Clouds Understanding} 
\begin{figure*}[htbp!]
      \centering
      \includegraphics[scale=0.1339]{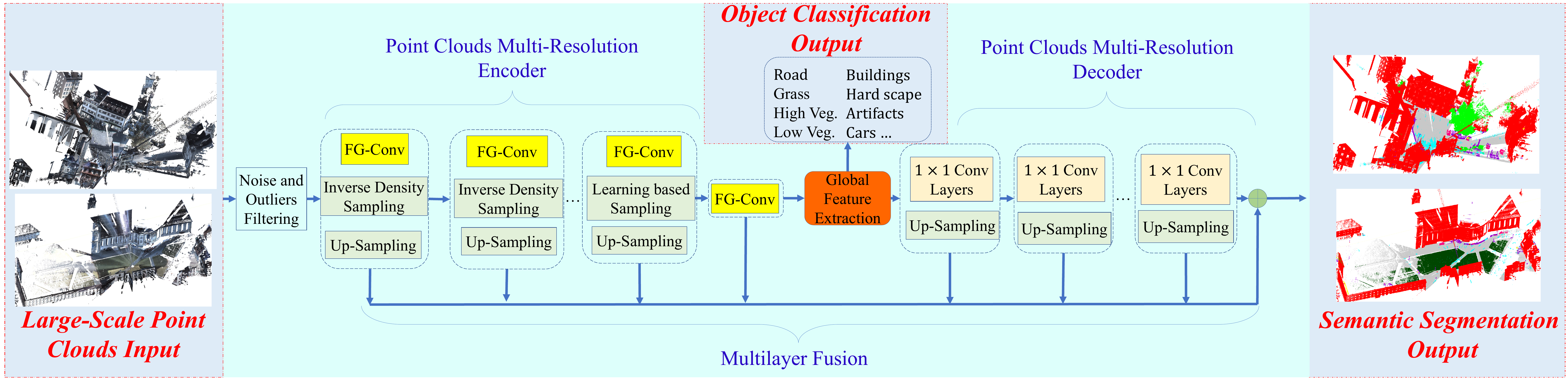}
      \caption{Overall system framework of proposed \textit{\textbf{FG-Net}}, and the core module \textit{\textbf{FG-Conv}} can be integrated into \textit{\textbf{FG-Net}} with multi-resolution residual learning. 1 $\times$ 1 Conv stands for 1 $\times$ 1 convolutions. The network operations are done from top to bottom and from left to right.}
      \label{fig_overall}
\end{figure*}
It is till recently that more attention has been paid to efficient large-scale point clouds understanding. Previously, block partitioning\cite{qi2017pointnet++}\cite{qi2017pointnet}\cite{wu2019pointconv} was utilized to divide large-scale point clouds into small $1m \times 1m$ sub-blocks before being fed to networks. However, such operation of partition is time-consuming and damages the spatial geometric contextual information among the objects of large-scale scene. 
Although several attempts \cite{landrieu2019point}  \cite{liu2019point} \cite{hu2020randla} have been made on large-scale point clouds segmentation, there are still some major problems existing: Firstly,  the farthest point sampling (FPS) adopted by most of the previous methods require large computational cost which increases quadratically\cite{wang2019re} with respect to the number of input points $N$. Secondly,  block partitioning causes that large-scale point clouds semantics can not be inferred within one scan, which limits the volume of point clouds that can be processed. Some methods \cite{liu2019point} \cite{shi2020pv} also try to combine voxel-wise features with pointwise features to improve the performance. Analogous to the super-pixel conception in 2D image domain, the super-point \cite{landrieu2019point} method in point clouds is also introduced to apply graph convolutions on large-scale points. But due to the high computational cost of voxelization or graph construction, such methods can hardly achieve real-time performance.

\section{Proposed Methodology} 


  
In this section, a fast deep learning method leveraging correlated feature mining and geometric-aware modelling is proposed for large-scale point clouds understanding.  
As illustrated in Fig. \ref{fig_overall}, our \textit{\textbf{FG-Net}} takes raw point clouds of a large-scale complex scene as input and gives the predictions of object classification and semantic segmentation simultaneously. The details are given as follows.
\subsection{Noise and Outliers Filtering}

\begin{algorithm}
     \caption{Radius and Gaussian Distribution Based Noise and Outliers Filtering} 
      \KwIn{The input raw point clouds $P^{in}=\{p_i\}, i=1, 2, ..., N^{in}$, $p_i=(x_i, y_i, z_i)$, where $N^{in}$ is the number of input points} 
      \KwOut{The points after filtering: $P^{out}=\{p_m\}, m=1, 2, ...,M^{out}$, $p_m=(x_m, y_m, z_m)$} 
        \For{all points $p_i \in P^{in}$ \textbf{in parallel}} 
        { 
            Find the radius $r$ neighbourhood of $p_i$, which is the ball $B_r=\{s \in \mathbb{R}^3, \|s-p_i\| \leq r$\}\; 
            Calculate the number of point $N^r$ in $B_r$, the points can be represented as: $\{p_j\},j={1,2,..., N^r}$\;
            \If{The number of points $N^r \leq \Omega$} 
                {Remove $p_i$ from $P^{in}$\;
                 $N^{in}=N^{in}-1$\;
                \textbf{Continue}}
             \For{j=1 to $N^r$ \textbf{in parallel}}
             {
             Calculate the distance ${d_{ij} = \sum\limits_{j} \|p_i-p_j\|}$\; 
             Model the distances by the Gaussian distribution, which means d $\sim N(\mu, \sigma)$,
             
             \If{$\sum_{j=1}^{N^r}d_{ij} \geq \mu + m\sigma$ or $\sum_{j=1}^{N^r}d_{ij} \leq \mu - m\sigma$} 
             {Remove $p_{ij}$ from $P^{in}$;}
            } 
            $P^{out}= P^{in}$\; 
        } 
        return $P^{out}$\; 
\label{noise_filter}
\end{algorithm}
The point clouds obtained by the LiDAR sensors contain noise and outliers which are harmful to the following high-level processing, thus we propose a novel filtering method for pre-processing. As shown in Algorithm \ref{noise_filter}, the set of input point clouds is defined as $P^{in}=\{p_{i}\}, \, i= 1, 2, 3,..., \, N^{in}, \,  p_{i}=(x_i, y_i, z_i)$. The radius based nearest neighbour ball query is conducted to ensure the robustness to the density distribution variation of point clouds in sampling. Given a query point $p_i \in P^{in}$, we define the neighbouring points within radius $r$ as $ =\{P_{j}\,|\, \|p_j-p_i\|\leq r, j=1, \, 2,\, 3... ,\,N^r\}$. The number of points $N^r$ is obtained, which can be regarded as an estimation of point density in radius $r$, and will be reused in the following inverse density based sampling (IDS).  If $N^r \leq \Omega$ ($\Omega$ is a threshold depending on the density of points acquired by LiDAR), we regard the point as an isolated point and remove it from $P^{in}$. Next, the distance between points is modelled as the Gaussian distribution, and the points are removed if the mean distance is outside the confidence interval according to Gaussian distribution.  The mean $\mu$ and standard deviation $\sigma$ of the distances between points can be computed as follows:



 \begin{equation}
 \mu=\frac{1}{N^{in}N^r} \sum_{i=1}^{N^{in}} \sum_{j=1}^{N^r} d_{ij}
\vspace{-1.6mm}
 \end{equation}
\begin{equation} \sigma = \sqrt{\frac{1}{N^{in}N^r} \sum_{i=1}^{N^{in}} \sum_{j=1}^{N^r} (d_{ij}- \mu)^2} 
\vspace{-1.6mm}
\end{equation}
The noise and outliers can be removed very effectively with a speed of 0.61s per million ($10^6$) points. This simple but effective method can be utilized to remove noise and outliers of point clouds while enhancing the performance of point clouds understanding in the meanwhile. The implementation details for acceleration of our framework will be introduced in subsection \textit{C}.

 

\subsection{Proposed Network Architecture for Large-Scale Point Clouds Understanding}

	
		
			
			
	
	

\begin{figure*}[ht]
\centering
\includegraphics[scale=0.52]{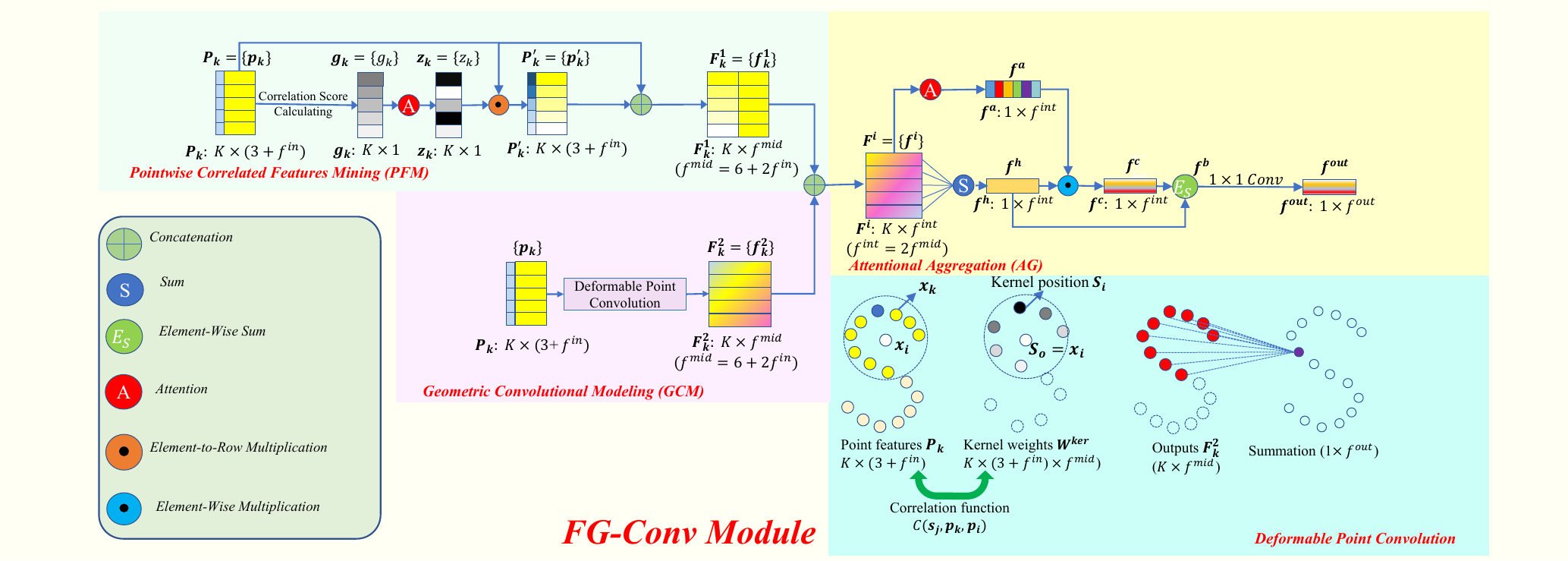}

\caption{Detailed illustration of our proposed novel pointwise correlated feature and geometric convolutional modelling module (\textbf{\textit{FG-Conv Module}}), the deformable convolution operation is illustrated at the right bottom corner, with the query point denoted as $p_i$ and the $\textit{k-th}$ neighbour point denoted as $p_k$, the output vector of point convolution is the dot product of point features and kernel weights. The feature vector is summed in attentional aggregation.}
\label{fig_FGConv}
\end{figure*}
We design the network module \textbf{\textit{FG-Conv}} to capture the feature correlations and model the local geometry of point clouds simultaneously. Leveraging feature pyramid based residual learning framework, \textbf{\textit{FG-Conv}} can be integrated into deep network \textit{\textbf{FG-Net}} as the core module for large-scale point clouds understanding. As shown in Fig. \ref{fig_FGConv}, the large-scale point clouds can be processed in parallel by our novel design leveraging feature-level correlation mining and geometric convolutional modelling. The core network module \textbf{\textit{FG-Conv}} includes 3 components: pointwise correlated features mining (PFM), geometric convolutional modelling (GCM), and attentional aggregation (AG), which are detailed as follows:
\subsubsection{Pointwise correlated features mining}

The point clouds after filtering are represented as x-y-z coordinates with per-point features. The input features can consist of raw RGB, surface normal information, intensity of point clouds, and even learnt latent features.
Denote the input point clouds as the matrix $\textit{\textbf{P}} \in \mathbb{R}^{N\times (3+f^{in})}$, where $N$ is the number of points and $f^{in}$ is the dimension of input features respectively. The \textit{i-th} vector in  $\textit{\textbf{P}}$ can be denoted as $\textit{\textbf{p}}_\textit{\textbf{i}}
=\textbf{(} \textit{\textbf{x}}_\textit{\textbf{i}},
\,
\textit{\textbf{f}}_\textit{\textbf{i}}
\textbf{)}
^{\mbox{\textit{\tiny{T}}}},$ where $\textit{\textbf{x}}_\textit{\textbf{i}} \in \mathbb{R}^3, \; \textit{\textbf{f}}_\textit{\textbf{i}} \in \mathbb{R}^{f^{in}},\ i=1, 2, 3, ..., N$.  For the point $\textit{\textbf{p}}_\textit{\textbf{i}}$,  denote the \textit{k-th} point vector in the spherical neighbourhood $B_r=\{\textit{\textbf{s}} \in \mathbb{R}^3, \|\textit{\textbf{s}}-\textit{\textbf{x}}_\textit{\textbf{i}}\| \leq r$\}\;  as $\textit{\textbf{p}}_\textit{\textbf{k}}=\textbf{(} \textit{\textbf{x}}_\textit{\textbf{k}}, \, \textit{\textbf{f}}_\textit{\textbf{k}}\textbf{)}^{\mbox{\textit{\tiny{T}}}}, \,
\textit{\textbf{x}}_\textit{\textbf{k}} \in \mathbb{R}^3,
\,
\textit{\textbf{f}}_\textit{\textbf{k}} \in \mathbb{R}^{f^{in}},\, k=1, 2, 3, ..., K$, in which $r$ is the radius of the neighbourhood. The similarity score $g_k$ of $\textit{\textbf{p}}_\textit{\textbf{k}}$ and $\textit{\textbf{p}}_\textit{\textbf{i}}$ is calculated as:

\begin{equation}
    g_{k}= \frac{\textit{\textbf{p}}_\textit{\textbf{k}}^{\mbox{\textit{\tiny{T}}}}
    \;
    \textit{\textbf{p}}_\textit{\textbf{i}}}{\|\textit{\textbf{p}}_\textit{\textbf{k}}\|\|\textit{\textbf{p}}_\textit{\textbf{i}}\|}
\end{equation}

which is the inner product of $\textit{\textbf{p}}_\textit{\textbf{k}}$ and $\textit{\textbf{p}}_\textit{\textbf{i}}$. It gives a good evaluation of the similarity of neighbouring points in spatial locations and features. For each of the \textit{K} neighbouring points, $g_{k}$ can be calculated and they constitute a similarity score vector $\textit{\textbf{r}}_\textit{\textbf{k}} \in \mathbb{R}^K, \, k=1, 2, 3, ..., K$. However, the similarity scores are not relevant to any specific task such as classification or segmentation. Thus, the attentional technique is introduced to make the new similarity score$\textit{\textbf{ z}}_\textit{\textbf{k}}$ adaptive to specific task by training of deep networks. which is calculated as:

\begin{equation}
    \textit{\textbf{z}}_\textit{\textbf{k}}=  \sigma (\textit{\textbf{w}}_\textit{\textbf{1}}\textit{\textbf{r}}_\textit{\textbf{k}}), \textit{\textbf{z}}_\textit{\textbf{k}} \in \mathbb{R}^K
\end{equation}
where $\textit{\textbf{w}}_\textit{\textbf{1}}\in \mathbb{R}^{K\times K}$ is the weight matrix to be learnt. $\sigma$ is the softmax function to normalize the attentional weights. All $\textit{\textbf{p}}_\textit{\textbf{k}}$ constitute the matrix $\textit{\textbf{P}}_\textit{\textbf{k}}=\textbf{(}\textit{\textbf{p}}_\textit{\textbf{1}},\, \textit{\textbf{p}}_\textit{\textbf{2}},\, \textit{\textbf{p}}_\textit{\textbf{3}},\, ...,\, \textit{\textbf{p}}_\textit{\textbf{k}}\textbf{)}^{\mbox{\textit{\tiny{T}}}}, \, \textit{\textbf{P}}_\textit{\textbf{k}}\in \mathbb{R}^{K\times (3+f^{in})}$. 
Then each element of $\textit{\textbf{z}}_\textit{\textbf{k}}$ is multiplied with each row of $\textit{\textbf{P}}_\textit{\textbf{k}}$ through element-to-row multiplication to obtain the augmented attentional feature matrix $\textit{\textbf{P}}{'}_k \in \mathbb{R}^{K\times (3+f^{in})}$. From now on, the augmented features (e.g.,   $\textit{\textbf{P}}{'}_k$) which encode both geometry and feature correlations are called feature for the sake of brevity. Next, $\textit{\textbf{P}}{'}_k$ is concatenated with their corresponding input feature $\textit{\textbf{P}}_\textit{\textbf{k}}$ to obtain the enhanced feature $\textit{\textbf{F}}_\textit{\textbf{k}}^\textit{\textbf{1}} \in \mathbb{R}^ {{K}{ \times f^{mid}}}$ $(f^{mid}=6+2f^{in})$. In this way, the local contextual relationship can be captured, and the similarity of features is enhanced adaptively and selectively by the attentive weighting in a learnable way. The similar feature elements in latent space are enhanced while distinct ones are attenuated.
\subsubsection{Geometric convolutional modelling}
After the pointwise correlated features mining, the local correlated features can be largely captured, but the geometric structure of points can not be sufficiently modelled.
Inspired by the great success of deformable convolution in the image recognition\cite{zhu2019deformable}, we extend deformable convolutions from image to point clouds to model the irregular and unordered 3D structures. Similar to 2D deformable convolutions, the deformable 3D kernels in Euclidean space are a set of learnable points that conform to the local structures of point clouds, thus, the dominant local geometric shapes of the points can be activated by the corresponding kernels in the neighbourhood. Note that kernel deformations can adapt to the local geometry of points in a learnable way by elaborately designed optimization functions.
As shown in the right bottom of Fig. \ref{fig_FGConv}, like convolutional neural networks in image processing, the convolution on points is defined as:
\begin{equation}
    \textit{\textbf{F}}_\textit{\textbf{k}}^\textit{\textbf{2}}\textit{\textbf{(}}\textit{\textbf{p}}_\textit{\textbf{k}}, \, \textit{\textbf{p}}_\textit{\textbf{i}}\,\textit{\textbf{)}} =\sum_ {\textit{\textbf{p}}_\textit{\textbf{k}} \in B_r} \textit{\textbf{K}}\,\textbf{(}\,\textit{\textbf{p}}_\textit{\textbf{k}}, \, \textit{\textbf{p}}_\textit{\textbf{i}}\textbf{)}\, \textit{\textbf{p}}_\textit{\textbf{k}}
\end{equation}
The core problem is that point clouds are unstructured and unordered, which makes it difficult for point convolutional kernel function $\,\textit{\textbf{K}}\,\textbf{(}\,\textit{\textbf{p}}_\textit{\textbf{k}}, \,\textit{\textbf{p}}_\textit{\textbf{i}}\,\textbf{)}$ to learn representative local geometric patterns. We design the correlation function to measure the correspondence between kernel points and local geometry. To be more specific, the closer kernels points are to input points, the higher the correlation value should be assigned. Denote the difference between $\textit{\textbf{x}}_\textit{\textbf{k}}$ and $\textit{\textbf{x}}_\textit{\textbf{i}}$ as $\Delta \textit{\textbf{x}}_\textit{\textbf{k}}= \textit{\textbf{x}}_\textit{\textbf{k}}-\textit{\textbf{x}}_\textit{\textbf{i}}$. The $N^s$ pseudo kernel points $\textit{\textbf{S}}_\textit{\textbf{i}}\subset B_r$  centered at $\textit{\textbf{S}}_\textit{\textbf{o}}$ ($\textit{\textbf{S}}_\textit{\textbf{o}}=\textit{\textbf{x}}_\textit{\textbf{i}}$) are designed so as to imitate the convolutional kernels in image processing, ($i={1, 2, 3, ..., N^s}$). The relative coordinates of pseudo kernel points $\textit{\textbf{S}}_\textit{\textbf{i}}$ and the center point $\textit{\textbf{S}}_\textit{\textbf{o}}$ are given as: $\textit{\textbf{s}}_\textit{\textbf{i}}=\textit{\textbf{S}}_\textit{\textbf{i}}-\textit{\textbf{S}}_\textit{\textbf{o}}$. We set the correlation function $C\textit{\textbf{(}}\textit{\textbf{s}}_\textit{\textbf{i}}, \textit{\textbf{p}}_\textit{\textbf{k}}, \textit{\textbf{p}}_\textit{\textbf{i}}\,\textit{\textbf{)}}$ as the Gaussian function formulated as:
\begin{equation}
    C\textit{\textbf{(}}\textit{\textbf{s}}_\textit{\textbf{i}}, \textit{\textbf{p}}_\textit{\textbf{k}}, \textit{\textbf{p}}_\textit{\textbf{i}}\,\textit{\textbf{)}} = \frac{1}{\|N^s\|}exp(-\frac{\|\textit{\textbf{s}}_\textit{\textbf{i}}-\Delta \textit{\textbf{x}}_\textit{\textbf{k}}\|^2}{m\sigma^2})
\end{equation}
where $N^s$ is the number of kernel points, $m$ is a constant, and $\sigma$ is the parameter determining the influence distance of kernel points. Then the kernel function can be given as the sum of all relations with learnable weights as shown in Equation \ref{kernel}:
\begin{equation}
\begin{aligned}
    K\textit{\textbf{(}}\textit{\textbf{p}}_\textit{\textbf{k}}, \textit{\textbf{p}}_\textit{\textbf{i}}\,\textit{\textbf{)}} &=\sum_{n=1}^{N^s} C\textit{\textbf{(}}\textit{\textbf{s}}_\textit{\textbf{i}}, \textit{\textbf{p}}_\textit{\textbf{k}}, \textit{\textbf{p}}_\textit{\textbf{i}}\,\textit{\textbf{)}}\,\textit{\textbf{W}}^{ker} \\&= \frac{1}{\|N^s\|} \sum_{n=1}^{N^s} exp(-\frac{\|\textit{\textbf{s}}_\textit{\textbf{i}}-\Delta \textit{\textbf{x}}_\textit{\textbf{k}}\|^2}{m\sigma^2})\, \textit{\textbf{W}}^{ker} \label{kernel} 
    \end{aligned}
\end{equation}
where $\textit{\textbf{W}}^{ker} \in \mathbb{R}^{(3+f^{in}) \times f^{mid}}$ is the weight matrix of MLP layers, and $3+f^{in}$, $ f^{mid}$ are input and output channel numbers respectively. During the optimization process, the kernel points are forced to adapt to the dominant structures in the local point clouds. Finally, the feature after deformable convolution $\textit{\textbf{F}}_\textit{\textbf{k}}^\textit{\textbf{2}} \in \mathbb{R}^ {{K}{ \times f^{mid}}}$ $(f^{mid}=6+2f^{in})$ can be obtained. In this way, the local geometric structures are well captured by convolutional kernels and the dominant structural features are enhanced.
\subsubsection{Attentional aggregation}
The attention mechanism is utilized to leverage the feature level and geometric level patterns without large information loss. As shown in Fig. \ref{fig_FGConv}, the integrated neighbouring features can be represented as $\textbf{\textit{F}}^\textbf{\textit{i}}\in \mathbb{R}^ {{K}{ \times f^{int}}}$ $(f^{int}=2f^{mid})$. Then the attentive score for aggregation is defined as $\textit{\textbf{w}}_\textit{\textbf{2}} \in \mathbb{R}^K$, which will adaptively learn the importance score of each feature. The weighted attentional feature $\textbf{\textit{f}}^\textbf{\textit{a}} \in \mathbb{R}^{f^{int}}$ can be given as:
\begin{equation}
\textbf{\textit{f}}^\textbf{\textit{a}} = softmax(\textit{\textbf{w}}_\textit{\textbf{2}} \, \textbf{\textit{f}}^\textbf{\textit{i}})
\end{equation}

The summed feature can be given as: $\textbf{\textit{f}}^\textbf{\textit{h}} = \sum_{i=1}^{K} \textbf{\textit{f}}^\textbf{\textit{i}}, \ \textbf{\textit{f}}^\textbf{\textit{h}} \in \mathbb{R}^{f^{int}}$. Then the element-wise multiplication between $\textbf{\textit{f}}^\textbf{\textit{a}}$ and $\textbf{\textit{f}}^\textbf{\textit{h}}$ is utilized to obtain the learnt feature $\textbf{\textit{f}}^\textbf{\textit{c}}= \textbf{\textit{f}}^\textbf{\textit{a}} \odot \textbf{\textit{f}}^\textbf{\textit{h}}$.
And the final features $\textbf{\textit{f}}^\textbf{\textit{b}}$ is the sum of original feature and learnt feature: $\textbf{\textit{f}}^\textbf{\textit{b}} = \textbf{\textit{f}}^\textbf{\textit{h}} + \textbf{\textit{f}}^\textbf{\textit{c}}$.
We apply the MLP layer to control the dimension of the output vector flexibly and give the meaningful aggregated feature $\textbf{\textit{f}}^\textbf{\textit{out}} \in \mathbb{R}^{f^{out}}$ containing both local correlated features and enhanced local geometry. 

\begin{figure*}[ht]
\centering
\includegraphics[scale=0.466]{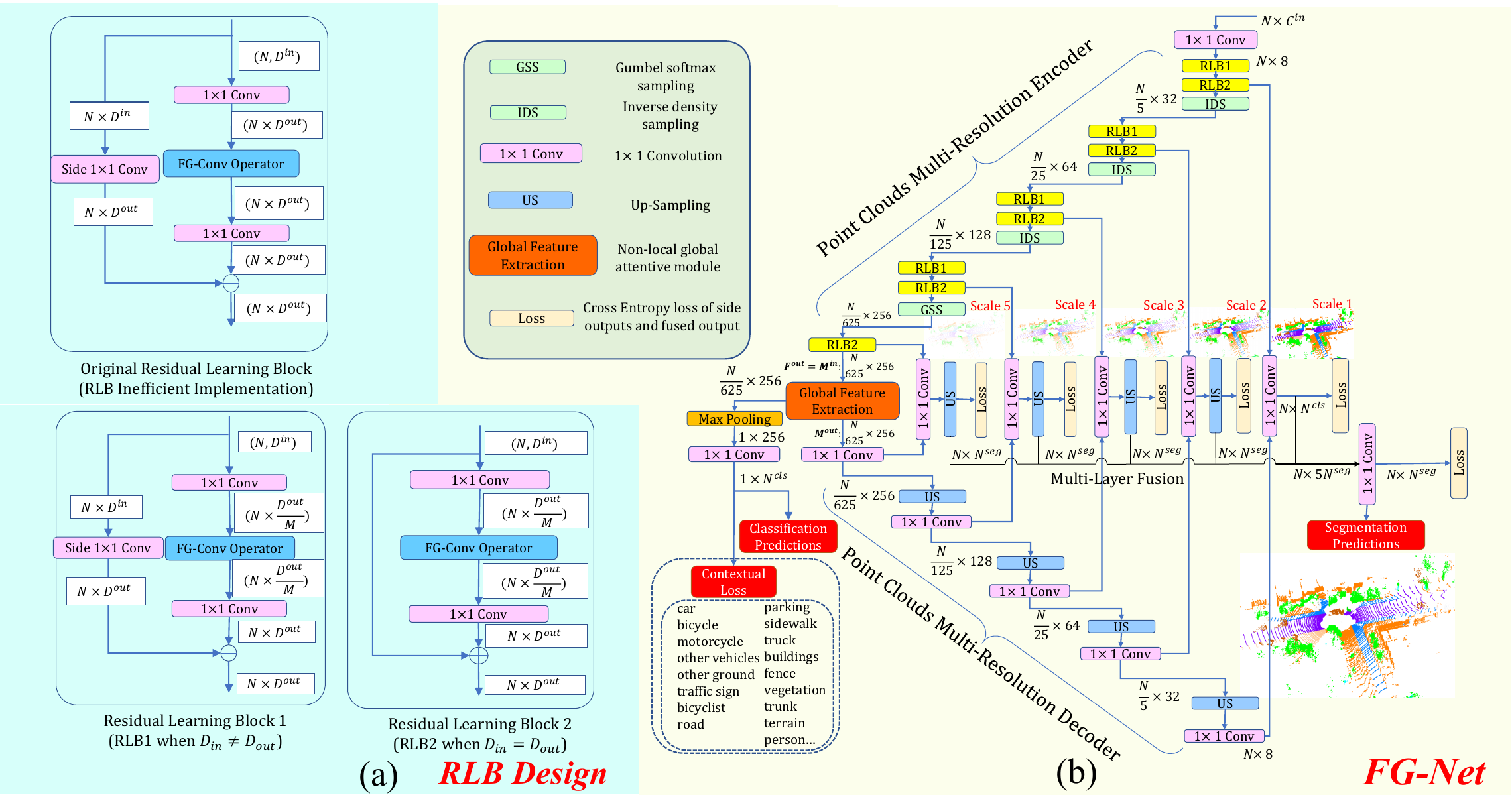}
\caption{The detailed residual learning block design of our $\textit{\textbf{FG-Net}}$ is shown in the subfigure (a), in which we show the original inefficient residual learning block and our designed residual learning block respectively, and the proposed feature pyramid residual learning network is shown in the subfigure (b). $N^{cls}$ and $N^{seg}$ stand for the number of classes in classification of the presence of objects or not, and number of classes in semantic segmentation respectively.}
\label{fig_resnet}
\end{figure*}

\begin{figure}[ht]
\centering
\includegraphics[scale=0.35]{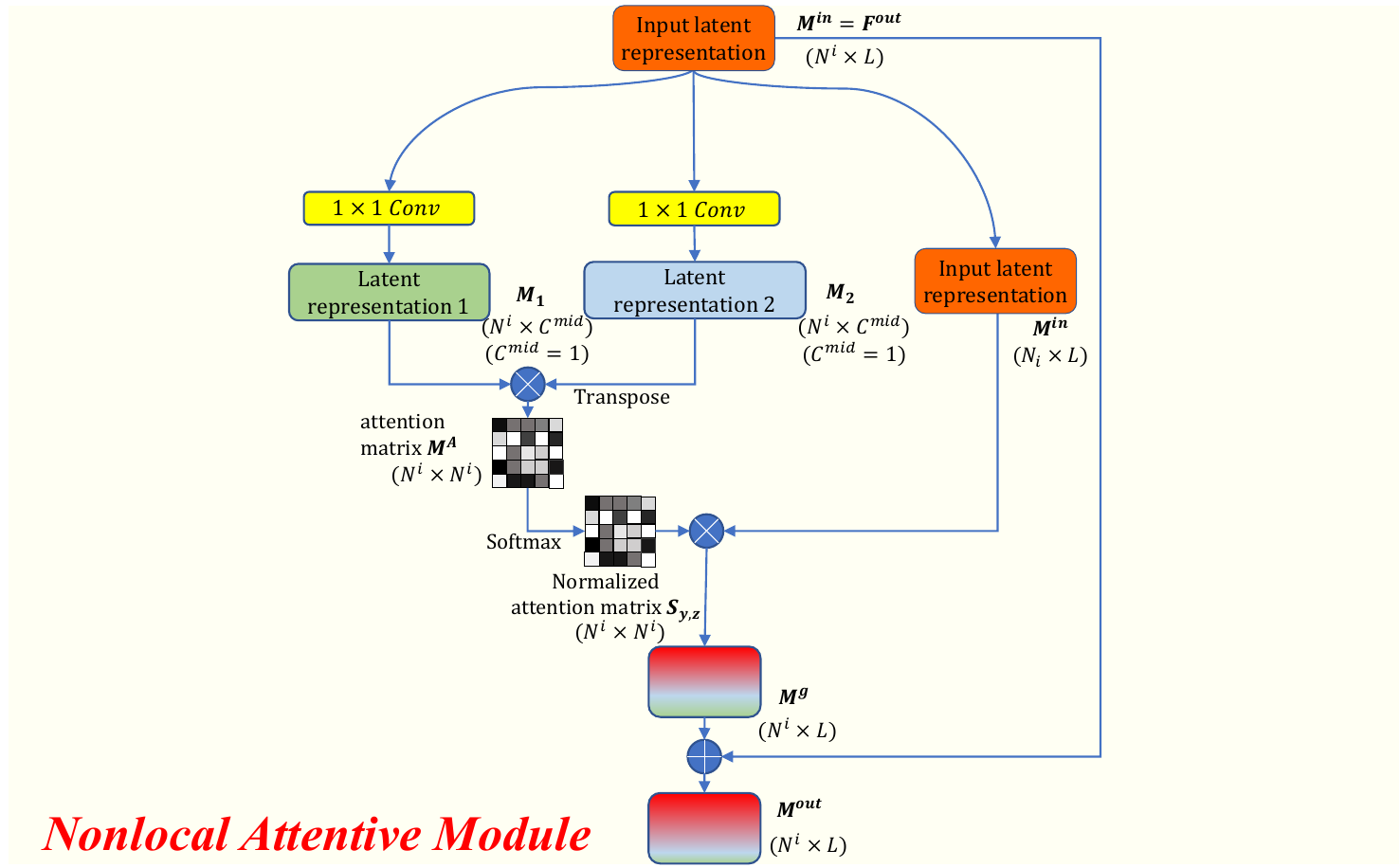}
\caption{Detailed illustration of the non-local global attentive module that functions as "Global feature Extraction" in Fig. \ref{fig_resnet}. The feature representations $\textbf{\textit{M}}^\textbf{\textit{in}}$ and $\textbf{\textit{M}}^\textbf{\textit{out}}$ in this module are corresponding to Subfigure (b) in Fig. \ref{fig_resnet}.}
\label{fig_nonlocal}
\vspace{-8.8mm}
\end{figure}
\subsubsection{Feature pyramid hierarchical residual architecture}

The Resnet \cite{He_2016_CVPR} based architecture has achieved great success in image recognition. Motivated by the residual learning paradigm, we proposed a general deep network that is specially designed for classification and semantic segmentation of large-scale point clouds. To our knowledge, until recently, there are several methods   \cite{liu2019densepoint} \cite{li2019deepgcns} starting to use residual learning for point clouds recognition, but their attempts are limited to small-scale point clouds. Thus the fitting capacity of residual architecture can not be fully demonstrated. We propose a feature pyramid based multi-scale fusion strategy for adaptively aggregating features from different layers of the network. Leveraging deep residual structure, memory-efficient deep networks can be built.

As shown in Fig. \ref{fig_resnet}, the encoder-decoder based network structure can be utilized to obtain point clouds at multiple resolutions. In image processing, the networks are supposed to extract large feature maps for small objects and small feature maps for big objects\cite{liu2020deep} \cite{wang2020deep}.  It should be noted that scale variation in images will not exist in 3D point clouds. Different from images in which the scale of objects will vary with the distance, the scale of point clouds will keep constant. Hence the deconvolution by interpolation must be conducted to recover the points to the original resolution. 
As shown in the Subfigure (a) of Fig. \ref{fig_resnet}, in residual learning block (RLB), denote the input dimension and output dimension of RLB as $D^{in}$ and $D^{out}$ respectively. Unlike some deep architectures which are memory consuming, we reduce the feature dimension in the original residual learning block to $D^{out} /{M}$ ($M=8$ is adopted in our framework) by $1 \times 1$ convolution before feeding them into \textbf{\textit{FG-Conv}} module, which reduce the parameters by 9.6 times. And the accuracy for classification and segmentation can also be maintained through residual learning, which will be given in experiments. Another $1 \times 1$ convolution will be applied to recover the feature dimension. At the block connecting two stages, $1\times 1$ convolution should be applied in skip link for increasing the feature dimensions. Then the global feature extraction in Subsubsection \textit{5} will be conducted to obtain the latent global features $\textbf{\textit{M}}^\textbf{\textit{out}}$ from $\textbf{\textit{M}}^\textbf{\textit{in}}$, which can be directly utilized for classification predictions. The point clouds are all upsampled after the $h$ ($h=5$ in our case) convolutional blocks. Unlike previous methods which directly used the upsampled features for segmentation, we propose to fuse the predictions at different resolutions and use the supervised loss to guide the training process. It turned out the hierarchical structure will give better results for pointwise large-scale point clouds segmentation. 


\subsubsection{Point clouds global feature extraction}
The global and long-range dependencies in point clouds should also be captured before doing upsampling and giving the pointwise predictions. Due to the limited receptive field of the neural layer mentioned above, the global contextual semantic patterns can not be fully obtained. We adopt the self-attentional module shown in Fig. \ref{fig_nonlocal} to selectively enhance the closely relevant elements in the global feature $\textbf{\textit{M}}^\textbf{\textit{in}}$. After the global relationship mining by this module, both the local and global relationships in features and geometry will be captured adaptively. Then the feature representation with combined local or non-local semantic contextual correlations will be adaptively obtained to facilitate the subsequent recognition task.  Given the original local feature map $\textbf{\textit{F}}^\textbf{\textit{out}}= \textbf{\textit{M}}^\textbf{\textit{in}} \in \mathbb{R}^{N^i \times L}$, $ N^i=\frac{N}{625}, L=256$ in our case) as shown in Fig. \ref{fig_resnet} and Fig. \ref{fig_nonlocal},  the $1 \times 1$ convolution with weight $\textbf{\textit{W}}^\textbf{\textit{G}} \in L \times C^{mid} \, (C^{mid}=1$ in our case) is used to transform the feature map into latent representations $\textbf{\textit{M}}_\textbf{\textit{1}}$ and $\textbf{\textit{M}}_\textbf{\textit{2}}$ for further obtaining the similarity of each two elements in $\textbf{\textit{F}}^\textbf{\textit{out}}$. After $\textbf{\textit{M}}_\textbf{\textit{1}}$ and $\textbf{\textit{M}}_\textbf{\textit{2}}$ are obtained, the dot product between them can be conducted to obtain the relevant score matrix $\textbf{\textit{M}}^{\mbox{\textit{\tiny{A}}}} \in \mathbb{R}^{N^i \times N^i}$ which is given as: 
\begin{equation}
    \textbf{\textit{M}}^{\mbox{\textbf{\textit{\tiny{A}}}}}= \textbf{\textit{M}}_\textbf{\textit{1}}  \textbf{\textit{M}}_\textbf{\textit{2}}^{\mbox{\textit{\tiny{T}}}}
\end{equation}

Each element $m_{y, z}$ in $\textbf{\textit{M}}^{\mbox{\textbf{\textit{\tiny{A}}}}}$ gives the relevance score between the representation $\textbf{\textit{M}}_\textbf{\textit{1}}$ and $\textbf{\textit{M}}_\textbf{\textit{2}}$. Then the softmax is applied to normalize the latent attentional scores to obtain the final self-relation weights $\textbf{\textit{S}}_\textbf{\textit{y, z}} \in \mathbb{R}^{N^i \times N^i}$ of the latent representation $M^{in}$. Each element $s_{y, z}$ of $\textbf{\textit{S}}_\textbf{\textit{y, z}}$ can be represented as:
\begin{equation}
    s_{y, z}= \frac{exp(m_{y, z})}{\sum_{y=1}^{N^i}\sum_{z=1}^{N^i}exp(m_{y, z})}
\end{equation}

The attention weights $s_{y, z}$ reveals the correlations among all local and non-local features. The more related distributed feature relationships, even in the non-local region, can be effectively captured, and larger attention weights are assigned in $s_{y, z}$ to enhances their similar semantic contexts. Finally, the attention scores are applied to all elements in $\textbf{\textit{M}}^\textbf{\textit{in}}$ to produce the global attentional vector $\textbf{\textit{M}}^\textbf{\textit{g}}$: 

\begin{equation}
    \textbf{\textit{M}}^\textbf{\textit{g}}= \textbf{\textit{S}}_\textbf{\textit{y, z}}  \textbf{\textit{M}}^\textbf{\textit{in}}
\end{equation}
and the consolidated feature $\textbf{\textit{M}}^\textbf{\textit{out}}$ is the sum of $\textbf{\textit{M}}^\textbf{\textit{g}}$ and $\textbf{\textit{M}}^\textbf{\textit{in}}$, which is given as: $\textbf{\textit{M}}^\textbf{\textit{out}}= \textbf{\textit{M}}^\textbf{\textit{g}} + \textbf{\textit{M}}^\textbf{\textit{in}}$.

Ultimately, the global contextual representation $\textbf{\textit{M}}^\textbf{\textit{g}}$ is fused with the local aggregated representation $\textbf{\textit{M}}^\textbf{\textit{in}}$ for a comprehensive encoding of local and non-local correlated features. The predictions of classification can be directly obtained from the aggregated latent features $\textbf{\textit{M}}^\textbf{\textit{out}}$ and the segmentation results can also be learnt by up-sampling. 

\subsubsection{Optimization function formulation and data augmentation}
As mentioned in Subsubsection 2, denote the relative coordinates of kernel points as $\textit{\textbf{s}}_\textit{\textbf{i}}, \,i={1, ,2, ..., N^s}$ and the learnt deformation as $\Delta \textit{\textbf{s}}_\textit{\textbf{i}}$. The losses utilized for the deformable convolution are designed as:
\begin{equation}
    L_{fit}(\Delta s_i)= \sum_{i=1}^{N^s} {min}_{\Delta \textit{\textbf{s}}_\textit{\textbf{i}}}(\frac{\|\Delta \textit{\textbf{x}}_\textit{\textbf{k}}-(\textit{\textbf{s}}_\textit{\textbf{i}}+\Delta \textit{\textbf{s}}_\textit{\textbf{i}})\|}{m\sigma^2})^2
\end{equation}
which is utilized to match the kernel positions with local geometries of point clouds.
\begin{equation}
    L_{rep1}(\Delta \textit{\textbf{s}}_\textit{\textbf{i}})={min}_{\Delta \textit{\textbf{s}}_\textit{\textbf{i}}} \sum_{i=1}^{N^s} \sum_{j=1}^{N^s} \frac{1}{\|\textit{\textbf{s}}_\textit{\textbf{i}}+ \Delta \textit{\textbf{s}}_\textit{\textbf{i}}-\textit{\textbf{s}}_\textit{\textbf{j}} -\Delta \textit{\textbf{s}}_\textit{\textbf{j}}\|}
\end{equation}
which is the repulsive loss utilized to keep distance between different kernels.
\begin{equation}
    L_{rep2}(\Delta \textit{\textbf{s}}_\textit{\textbf{i}})={min}_{\Delta \textit{\textbf{s}}_\textit{\textbf{i}}}
    \sum_{i=1}^{N^s} {\|\textit{\textbf{s}}_\textit{\textbf{i}}+ \Delta \textit{\textbf{s}}_\textit{\textbf{i}}\|^2}
\end{equation}
which is to keep the kernel points from diverging and make them inside the query ball. The kernel loss will be the sum of above 3 losses. i.e. $L_{ker}(\Delta \textit{\textbf{s}}_\textit{\textbf{i}})=L_{fit}(\Delta \textit{\textbf{s}}_\textit{\textbf{i}})+L_{rep1}(\Delta \textit{\textbf{s}}_\textit{\textbf{i}})+ L_{rep2}(\Delta\textit{\textbf{s}}_\textit{\textbf{i}})$.
As shown in Fig. \ref{fig_resnet}, the losses at different stages of the network are also summed, which can be formulated as the cross-entropy loss denoted as $L_1$:
\begin{equation}
\begin{aligned}
    L_1(\textit{\textbf{W}}) =& \sum_{n=1}^{N} (\sum_{h=1}^{H} \alpha^{(h)} \hat{y}_i log(P_{seg}(\textit{\textbf{p}}_\textit{\textbf{i}}^{(h)}, \textit{\textbf{W}}\,)) \\& + \beta \hat{y}_i^{fuse} log(P_{seg}(\textit{\textbf{p}}_\textit{\textbf{i}}^{(fused)}, \textit{\textbf{W}}\,))
\end{aligned}
\end{equation}
$\alpha^{(h)}$ denotes the weight at stage $h$ of the residual network, $\textit{\textbf{W}}$ denotes the weight of the entire network, $\textit{\textbf{p}}_\textit{\textbf{i}}^{(h)}$ denotes the upsampled point clouds at stage $h$, $\textit{\textbf{p}}_\textit{\textbf{i}}^{(fused)}$ denotes the fused point clouds, $\hat{y}_i$ and $\hat{y}_i^{fuse}$ denote the segmentation ground truth of points at different stages and fused points respectively. And $P_{seg}$ denotes the segmentation prediction of the networks. We also propose to use the contextual loss shown in Fig. \ref{fig_resnet} to predict the presence of objects or not in the scene to consider semantic contexts of the scene, which can be given as:
\begin{equation}
\begin{aligned}
    L_2(\textit{\textbf{W}}) =& \sum_{i=1}^{I} \hat{y}^{pre}_i log(P_{cls}(\textit{\textbf{p}}_\textit{\textbf{i}}, \textit{\textbf{W}}\,))
\end{aligned}
\end{equation}
Where $\hat{y}^{pre}_i$ indicates whether the object presents in the scene or not and $P_{cls}$ is the classification prediction. This loss helps the network equally consider all the semantic categories appearing in the scene.
The total loss the network can be given as: $ L_c(\textit{\textbf{W}}, \, \Delta \textit{\textbf{s}}_\textit{\textbf{i}} )=L_1(\textit{\textbf{W}})+ L_2(\textit{\textbf{W}})+L_{ker}(\Delta \textit{\textbf{s}}_\textit{\textbf{i}})$. The kernel positions and network parameters are jointly optimized in an end-to-end manner. 


\begin{figure}[bp]
\centering
\includegraphics[scale=0.215]{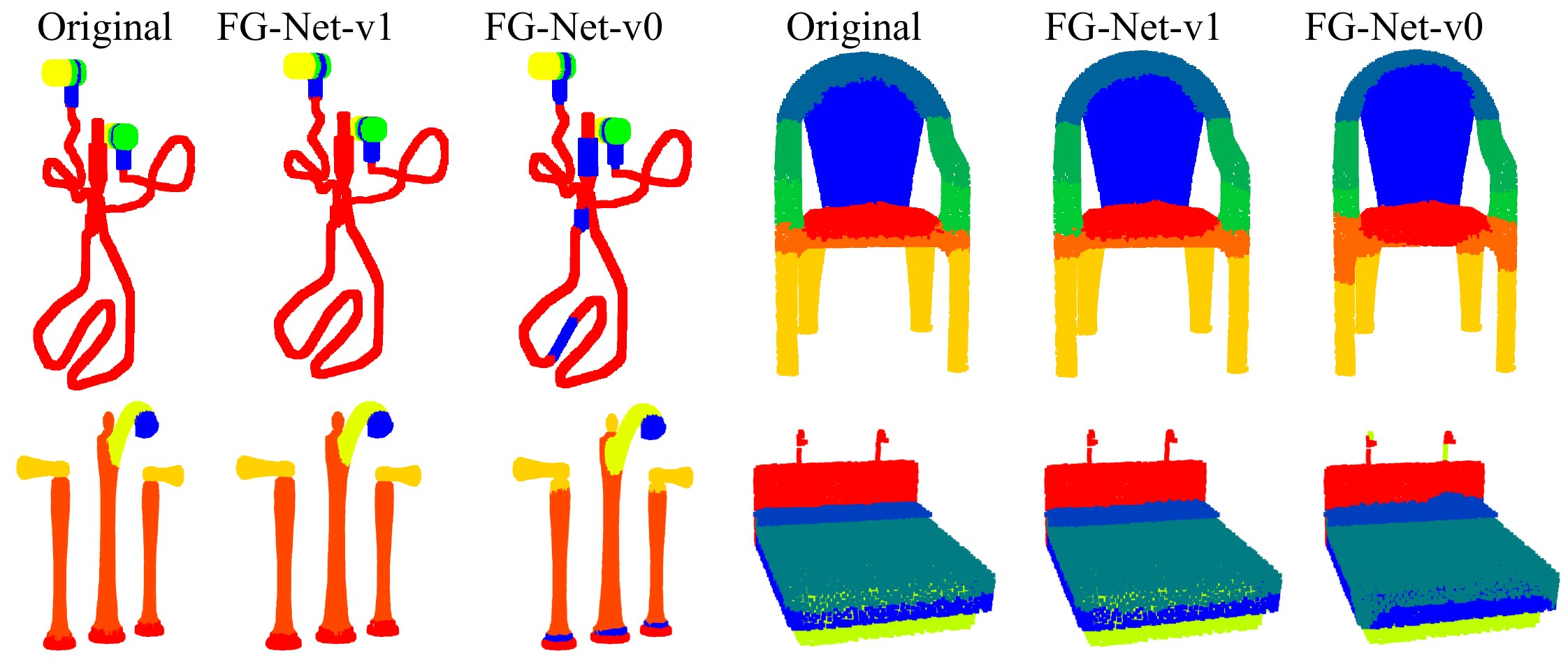}
\caption{Segmentation results of PartNet (\textit{\textbf{FG-Net-v0}} is the network with global attention while \textit{\textbf{FG-Net-v1}} is the one without global attention).}
\label{fig_partnet}
\end{figure}

\begin{figure*}[ht]
\centering
\includegraphics[scale=0.26]{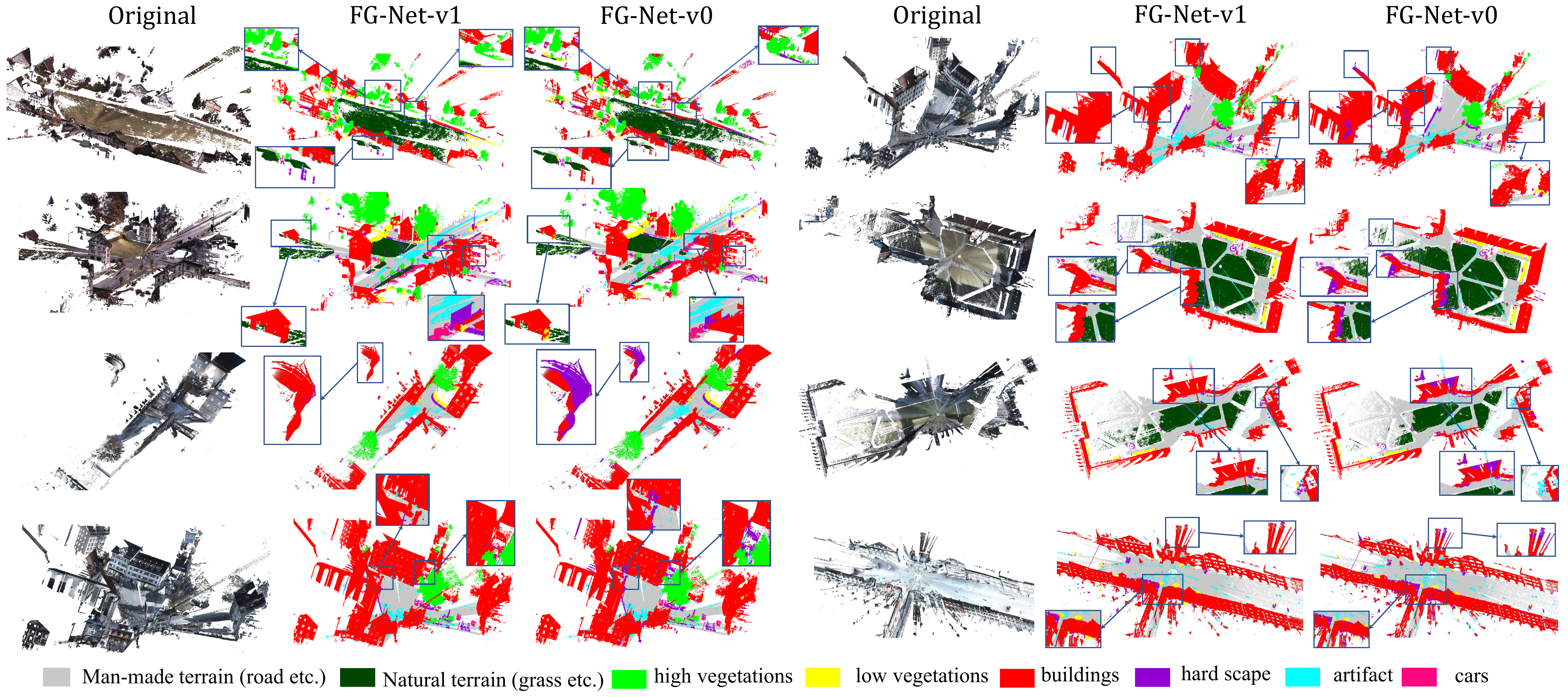}
\caption{Semantic3D segmentation results with the inconsistent predictions indicated by the black circles, please zoom in for details (\textit{\textbf{FG-Net-v0}} is network with global attention, while \textit{\textbf{FG-Net-v1}} is the one without global attention).}
\label{fig_Sem3D}
\end{figure*}

\begin{figure*}[ht]
\centering
\includegraphics[scale=0.32]{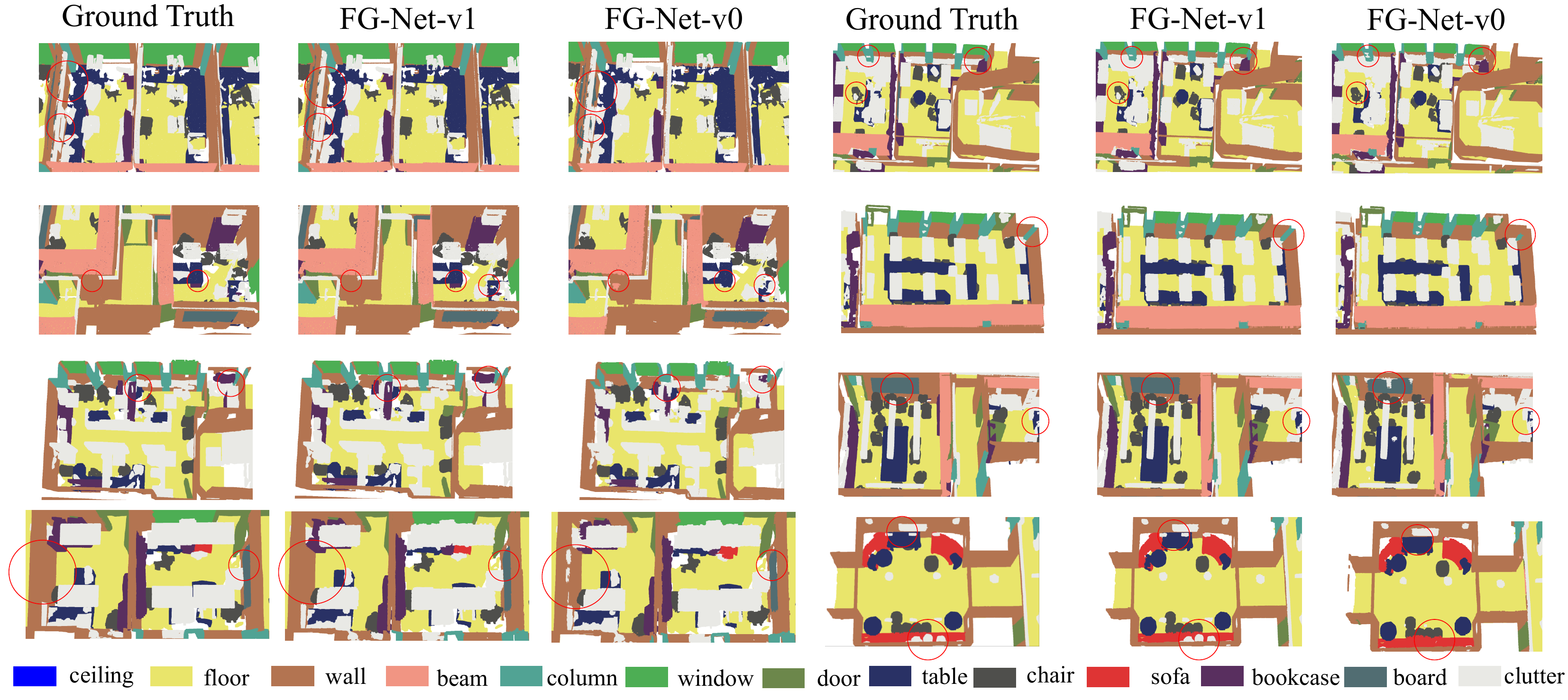}
\caption{Detailed S3DIS segmentation results with the inconsistent predictions highlighted in the red circles.}
\label{fig_s3dis_detail}
\end{figure*}

\begin{figure*}[htbp!]
\centering
\includegraphics[scale=0.323]{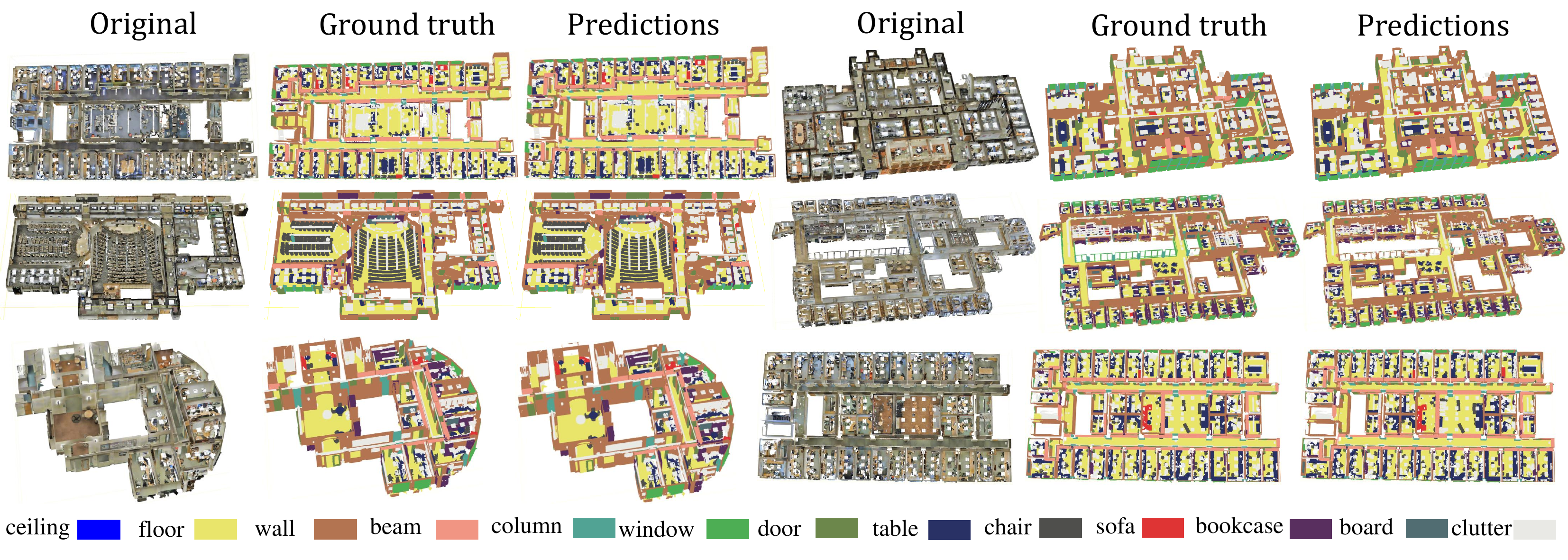}
\caption{S3DIS segmentation results of the whole area of Area 1 to Area 6, please zoom in for details.}
\label{fig_s3dis}
\vspace{-3.99mm}
\end{figure*}


\begin{figure}[ht]
\centering
\includegraphics[scale=0.26]{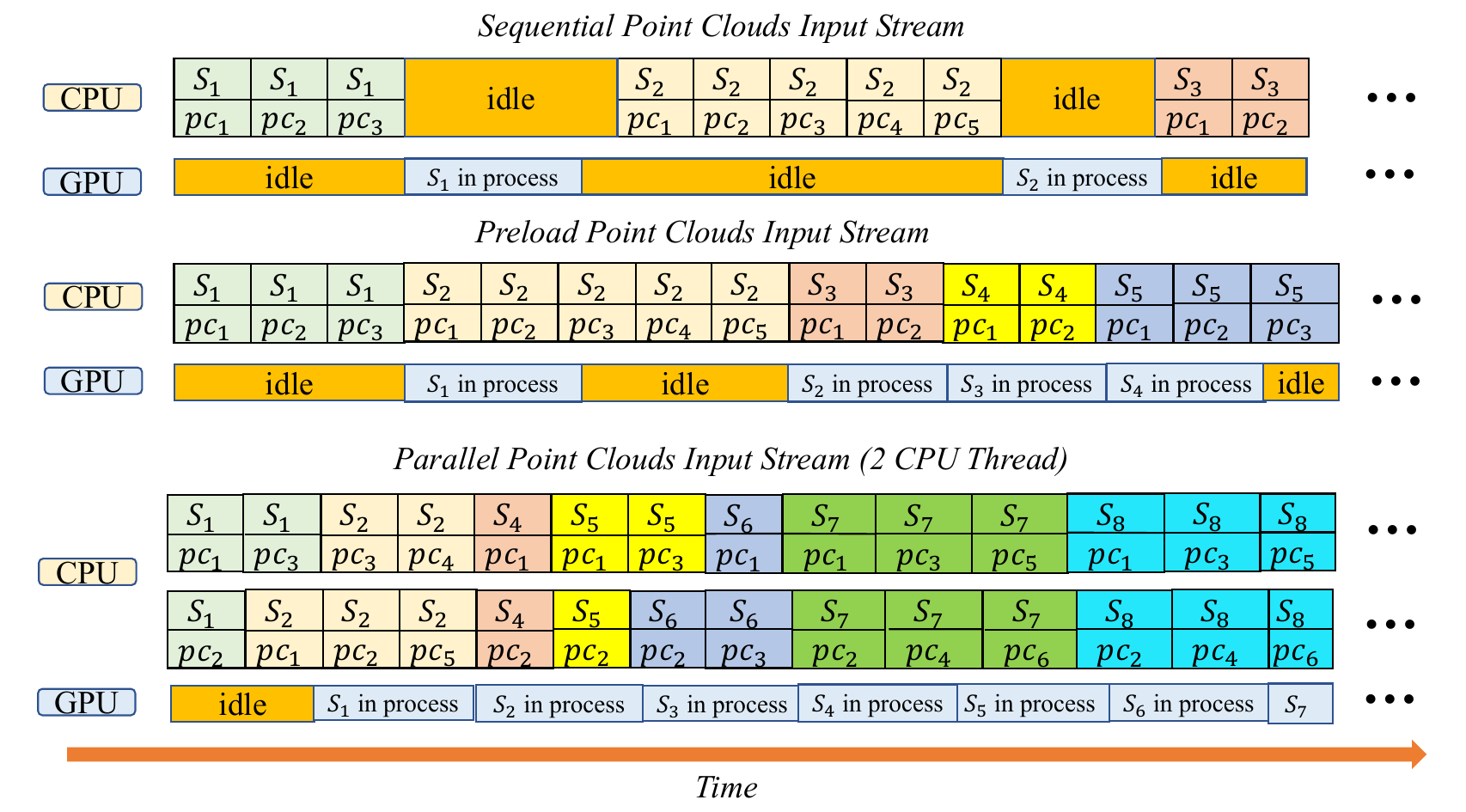}
\caption{The multi-thread CPU parallel processing of point clouds streams. The noise and outliers filtering is done on the CPU while sampling and deep network processing are done on the GPU. $S_i$ stands for $\textit{i-th}$ CPU processing stream, $pc_j$ stands for $\textit{j-th}$ point clouds batch in a single CPU stream.}
\label{fig_para}
\vspace{-6.00mm}
\end{figure}

\begin{figure*}[htbp!]
\centering
\includegraphics[scale=0.23]{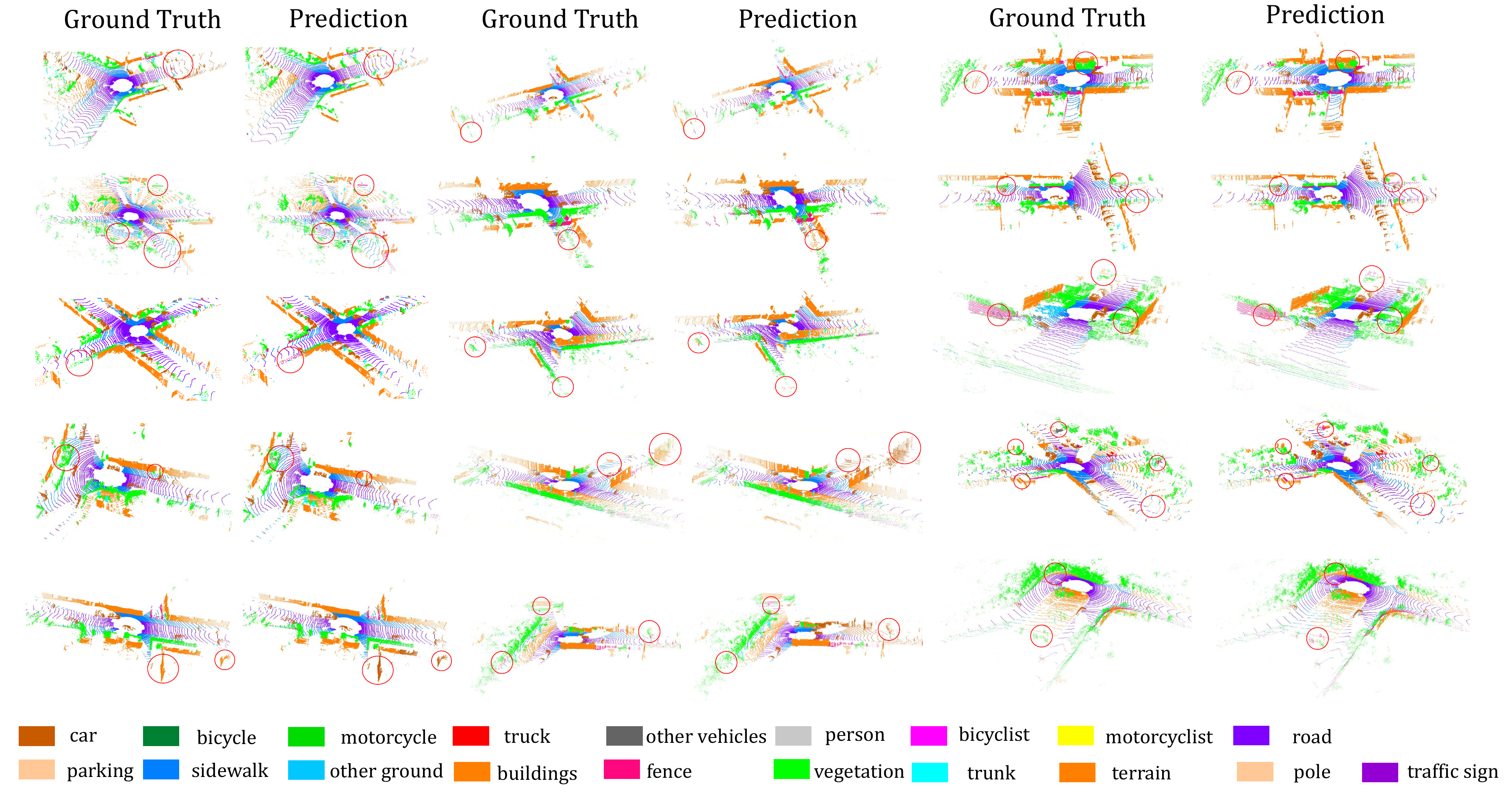}
\caption{Visualization of the Semantic-KITTI segmentation results with the wrong predictions highlighted by the red circles, please zoom in for details}
\label{fig_Skitti}
\end{figure*}

\subsection{Implementation Details for Acceleration of our Framework}

\subsubsection{\textit{\textbf{IGSAM}} for fast learning-based sampling}
The sampling methods play a very significant role in processing point clouds by convolutional neural networks.  Directly using raw points for segmentation is rather inefficient due to large computational costs when feeding them all into networks. Thus, an effective sampling method is highly required when taking efficiency into consideration. To tackle the large computational overhead when processing millions of point clouds, we propose efficient sampling methods \textit{\textbf{IGSAM}} leveraging the advantages of inverse density sampling (IDS) and gumbel softmax sampling (GSS) to achieve fast and effective point clouds understanding. We design a novel learning based GSS which adaptively selects the significant points based on optimization objectives. Leveraging inverse density sampling (IDS), point clouds can be sampled efficiently with density awareness while the meaningful information or feature for point clouds understanding is maintained.

To achieve learning based sampling operation, the Gumbel-Softmax trick is utilized to transfer the non-differentiable sampling operation into differentiable selection of features and coordinates by reparameterization tricks \cite{maddison2016concrete}. In this way, points that matter most for the task can be selected in a learnable way. Given the point clouds $\textit{\textbf{P}} \in \mathbb{R}^{N\times (3+f^{in})}$, with its coordinate and features, the probability score $\textit{\textbf{s}} \in \mathbb{R}^{N}$ of a point being sampled can be estimated by the MLP based $1\times 1$ convolution, which is formulated as:
\begin{equation}
\textit{\textbf{s}}= softmax(MLP(\textit{\textbf{P}})) 
\label{eqmlp}
\end{equation}
 Then the gumbel noise \cite{maddison2016concrete} $\textit{\textbf{g}} \in \mathbb{R}^{N}, i = 1, 2 , 3, ..., N$ can be selected from the gumbel distribution $Gumbel (0, 1)$. And the \textit{gumbel softmax} (\textit{GS}) operation on $\textit{\textbf{P}}$ can be used to compute the sampled vector $\textit{\textbf{y}}_\textit{\textbf{i}}$, where $\tau$ is the time constant, and $\textit{\textbf{s}}_\textit{\textbf{i}}$ and $\textit{\textbf{g}}_\textit{\textbf{i}}$ are the \textit{i-th} element in $\textit{\textbf{s}}$ and $\textit{\textbf{g}}$ respectively:
\begin{equation}
    \textit{\textbf{y}}_\textit{\textbf{i}}=GS(\textit{\textbf{s}}_\textit{\textbf{i}})= softmax((log(\textit{\textbf{s}}_\textit{\textbf{i}})+\textit{\textbf{g}}_\textit{\textbf{i}})/\tau) \label{select}
\end{equation}
Utilizing Equation \ref{select}, $\textit{\textbf{y}}_\textit{\textbf{i}}$ is differentiable with respect to $\textit{\textbf{s}}_\textit{\textbf{i}}$, thus, the differentiable sampling can be realized for original point distributions in the training process. It should be noted that $\tau$ should start from a high value of 1.0 and gradually anneals to a smaller value of 0.05 during training.
 When $\tau \to 0$, \textit{GS} degenerates to the Gumbel-Max (\textit{GM}) selection. And the discrete individual point candidate can be hard selected utilizing \textit{GM} in inference as shown in Equation \ref{one_hot_encode}:
 \begin{equation}
GM(\textit{\textbf{s}}_\textit{\textbf{i}})= one\_hot\_encode(argmax(log(\textit{\textbf{s}}_\textit{\textbf{i}})+\textit{\textbf{g}})) \label{one_hot_encode}
\end{equation}
 In this way, we can facilitate the differentiable sampling in training while maintain a discrete and hard \textit{GM} sampling in network inference, hence sampling operation can be integrated into deep networks.  
To be more specific, given the input points denoted as $\textit{\textbf{P}} \in R^{N \times (d+3)}$, the \textit{GSS} can be conducted as:

 \begin{equation}
     \textit{\textbf{P}}_s=  GS(\textit{\textbf{W}}^s \textit{\textbf{P}}{^{\mbox{\textit{\tiny{T}}}}}) \cdot \textit{\textbf{P}} 
     \label{output}
 \end{equation}
 where $\textit{\textbf{W}}^s \in \mathbb{R}^{N^{s} \times (d+3)}$ is weights of MLP layers in Equation \ref{eqmlp}, and $\textit{\textbf{P}}_s \in \mathbb{R}^{N^{s} \times (d+3)}$ are the $N^s$ points to be selected.

During the inference of the network, the discrete sampling can be realized by substitute \textit{GS} with \textit{GM}, which is formulated as:

 \begin{equation}
     \textit{\textbf{P}}_s=  GM(\textit{\textbf{W}}^s \textit{\textbf{P}}{^{\mbox{\textit{\tiny{T}}}}}) \cdot \textit{\textbf{P}}
 \end{equation}
Note that the matrix $\textit{\textbf{W}}^s$ will be relatively memory consuming when the quantity of point clouds is large, hence, we only adopt it in deeper layers of network when the points are down sampled to a certain quantity, which is less than $10^5$.

In our \textit{\textbf{IGSAM}}, the IDS is conducted in the first four stages of the network while the GSS is adopted in the fifth layer of the network. In this way, the density awareness can be maintained while significant points can be selected in a learnable way before conducting local aggregation and mining the global relationship by the non-local attentional module as introduced above, which enforces the sampling process to select the significant points for the specific understanding task.
\subsubsection{Acceleration by multi-thread parallel computation on CPU}

The noise and outliers filtering is implemented on the CPU while deep network computations are run on the GPU. It should be noted that we have reused the radius based ball query in both point clouds filtering and network operations for accelerations. As shown in Fig. \ref{fig_para}, we have preloaded the next stream of point clouds to CPU before the network computation on GPU is finished for acceleration. And the multi-thread computation on CPU is also utilized to accelerate the query process which reduces the idle period notably in subsequent CPU and GPU computations.


	
	
		
		
		
		 
		
		
		 
\vspace{-1.06mm}
\section{Experiments Results}
\subsection{Experiments Setup}
We have tested our method extensively on 8 different large-scale point clouds understanding datasets. We implemented the network in Tensorflow and optimized it with Adam optimizer and initial learning rate of $1e^{-4}$.
Also, the point clouds are randomly rotated around each axis $x, y, z$ with an angle $\phi \in [0, 2\pi]$. The scaling is also applied along $x, y, z$ axis with a scalar $\mu \in [0.85, 1.15]$ for data augmentation. The network is trained and tested in parallel with $5 \times 10^5$ point clouds in each stream. The experiments are conducted on Nvidia GTX 1080 graphics card with 8 GB memory. 
\subsection{Experiments of Filtering and Sampling Methods}

\begin{figure}[htbp!]
\centering
\includegraphics[scale=0.26]{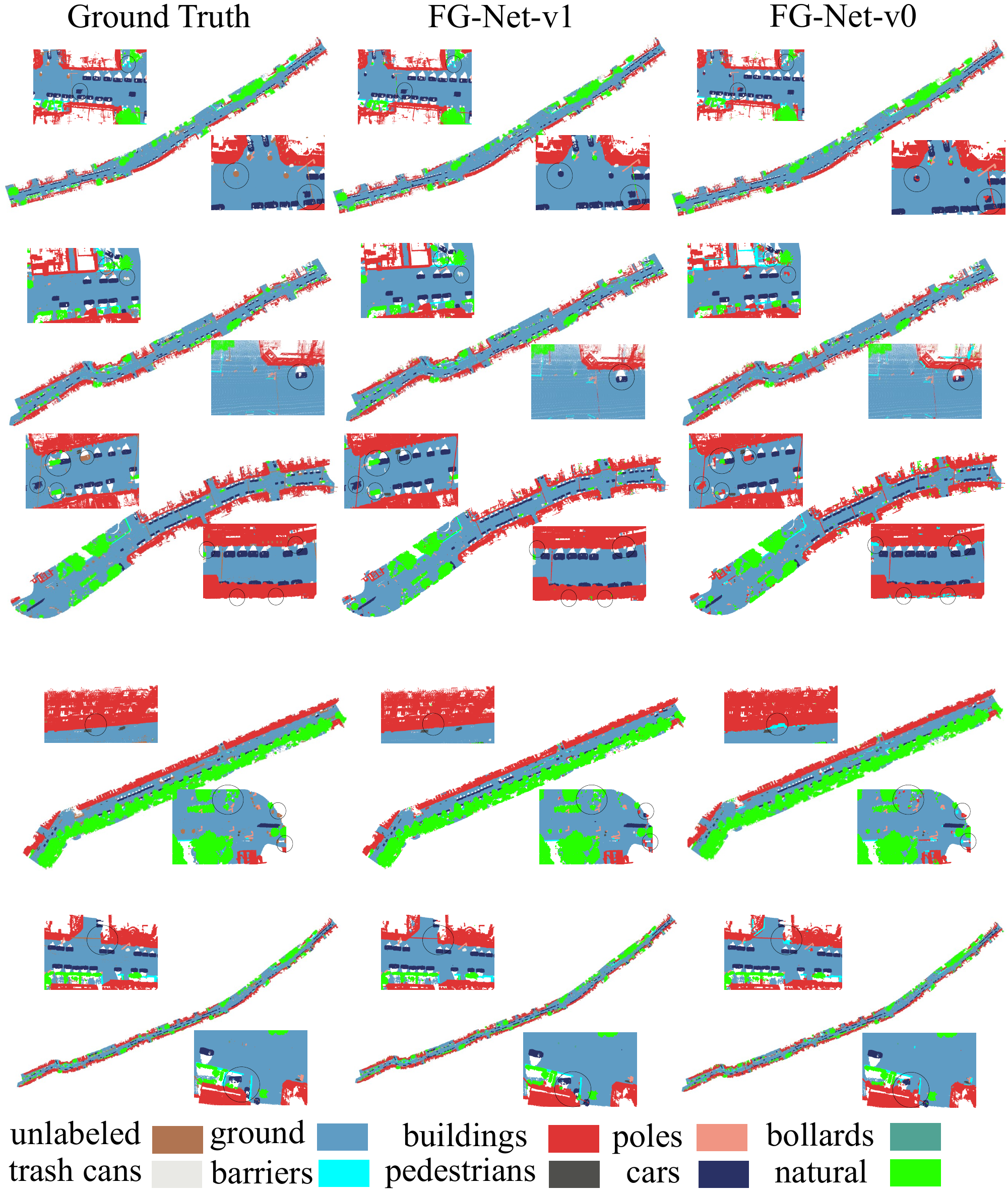}
\caption{NPM3D segmentation results, it should be noted that the color brown stands for unlabeled points, please zoom in for details.}
\label{fig_NPM3D}
\end{figure}
\begin{figure}[htbp!]
\centering
\includegraphics[scale=0.120]{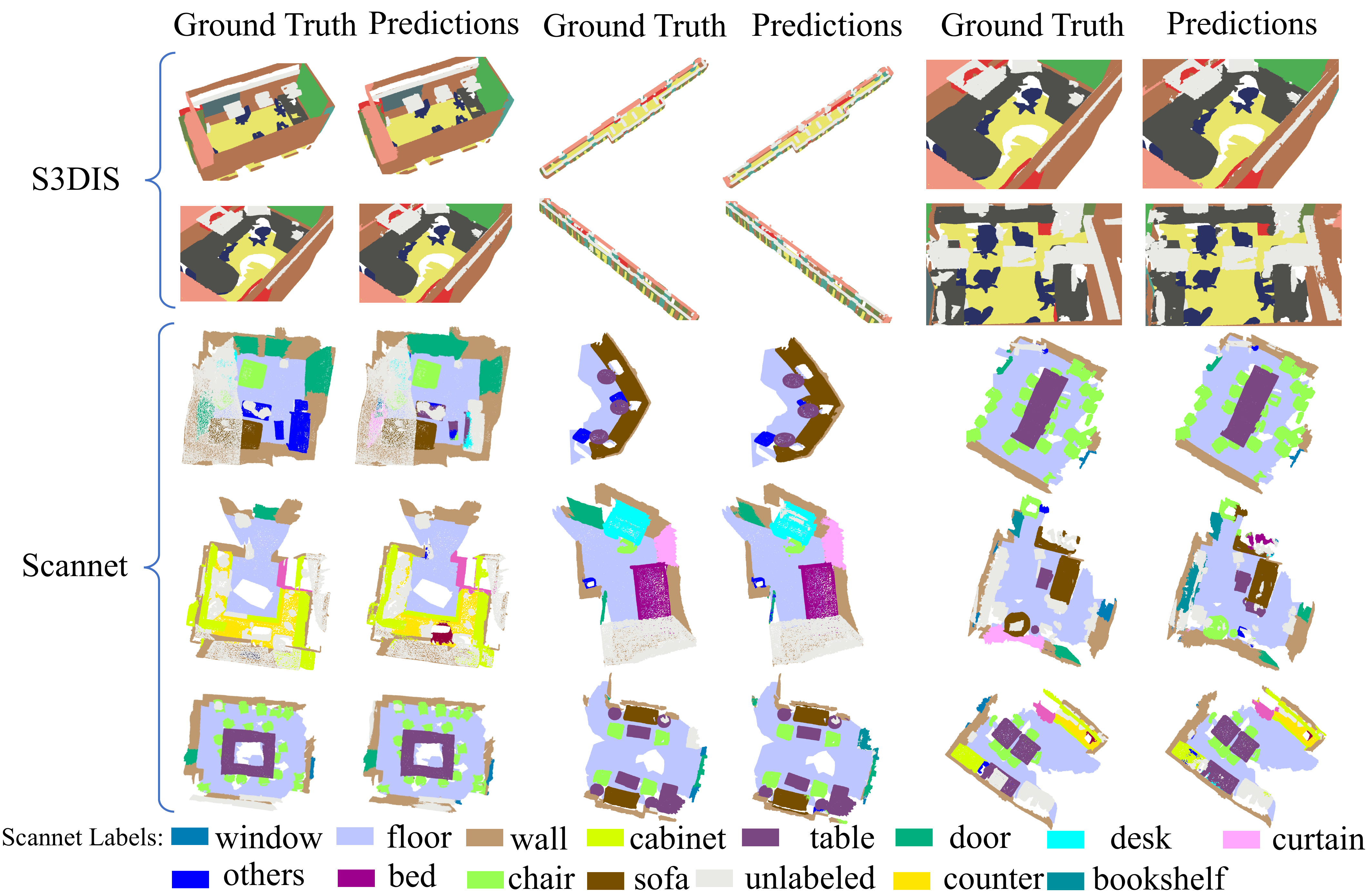}
\caption{Transfer learning results between S3DIS and Scannet, please zoom in for details}
\label{fig_transfer}
\end{figure}
\subsubsection{Noise and Outliers filtering}
We have tested the influence of proposed noise and outliers filtering on the semantic segmentation performance of S3DIS. Noted that we adopt 6-fold cross-validation for S3DIS to guarantee the generality and robustness. The noise filtering results with mean intersection over unions (mIOUs) are shown in Table \ref{table_filter}, it demonstrates that filtering has a boost on segmentation performance for diverse point clouds understanding methods. With unrelated isolated noise points removed, the meaningful semantics of point clouds will be retained which will boost the performance of segmentation.
\begin{table}[bp!]
\caption{The comparison of segmentation on S3DIS with and without filtering.}
\label{table_filter}
\begin{center}
\begin{tabular}{ccc}
\toprule

Method & mIOUs (without) & mIOUs (with)\\

\hline
SPG\cite{landrieu2019point}&62.1&63.2\\ 

Shellnet\cite{zhang2019shellnet}&66.8&67.9\\

PointCNN\cite{wu2019pointconv}&65.4&65.8\\

Kpconv\cite{thomas2019kpconv}&67.1&68.5\\

DGCNN\cite{wang2019dynamic}&56.1&58.2\\
FKA-Conv\cite{boulch2020fka}&68.1&68.6\\
\textbf{\textit{FG-Net}} (Ours) \textbf{\textit{(best)}}&\textbf{70.2}&\textbf{70.8}\\
\bottomrule
\end{tabular}
\end{center}
\vspace{-2mm}
\end{table}
\subsubsection{Point clouds sampling methods}
\begin{figure}[htbp!]
\centering
\includegraphics[scale=0.320]{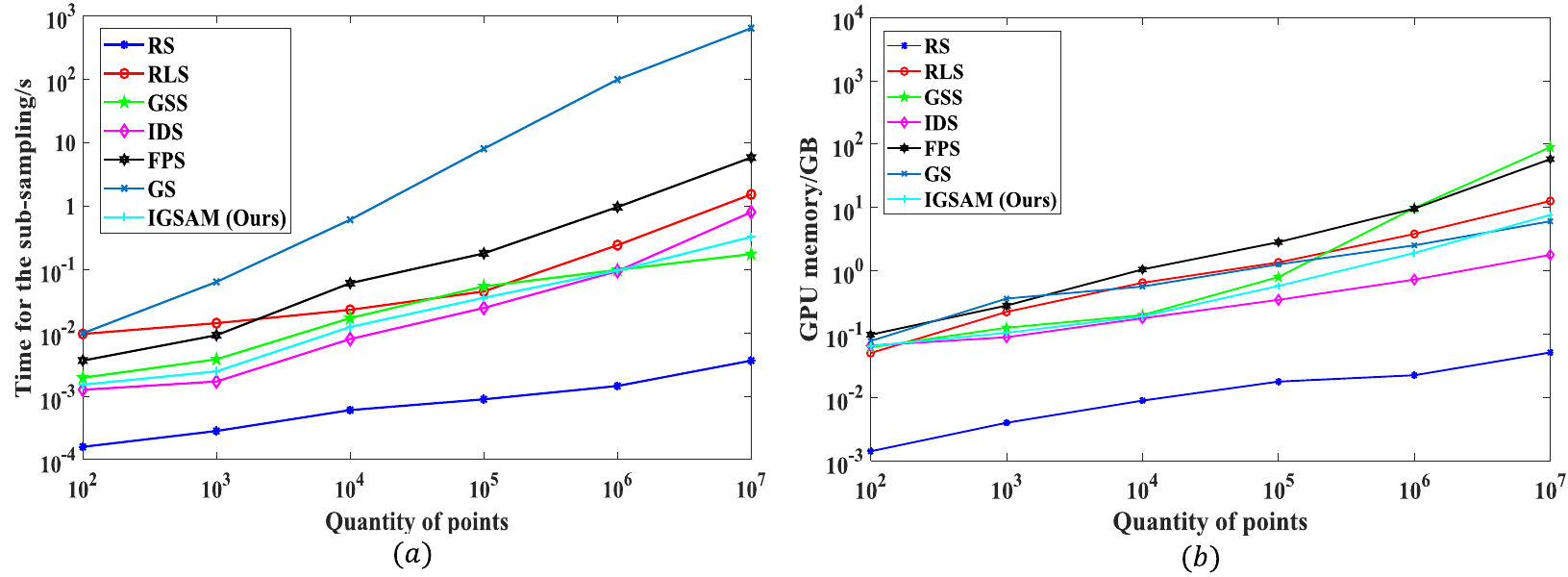}
\caption{Comparison of computational cost (a) and memory consumption (b) of different sampling methods.}
\label{fig_comp}
\vspace{-7mm}
\end{figure}
\begin{table}[htbp!]

\caption{The comparison of segmentation performance on S3DIS with diverse sampling methods}
\label{table_sampling}
\begin{center}
\begin{tabular}{ccccccc}
\toprule

Sampling Method & RS & RLS & \textit{\textbf{IGSAM}} & IDS & FPS & GS\\

\hline
SPG\cite{landrieu2019point}&61.7&62.2&63.2&62.6&61.8&62.5\\ 

Shellnet\cite{zhang2019shellnet}&66.2&66.3&67.6&66.9&66.1&66.3\\

PointCNN\cite{wu2019pointconv}&64.9&65.3&66.8&66.1&65.9&65.3\\

Kpconv\cite{thomas2019kpconv}&66.5&67.1&67.5&67.3&66.7&66.3\\

DGCNN\cite{wang2019dynamic}&55.2&56.0&58.4&57.5&57.9&56.0\\
FKA-Conv\cite{boulch2020fka}&64.6&65.2&68.6&66.2&66.1&65.3\\
\textbf{\textit{FG-Net}} (Ours) \textbf{\textit{(best)}}&\textbf{66.8}&\textbf{70.3}&\textbf{70.8}&\textbf{70.3}&\textbf{69.5}&\textbf{69.9}\\
\bottomrule
\end{tabular}
\end{center}
\end{table}


 \begin{figure}[ht]
\centering
\includegraphics[scale=0.21]{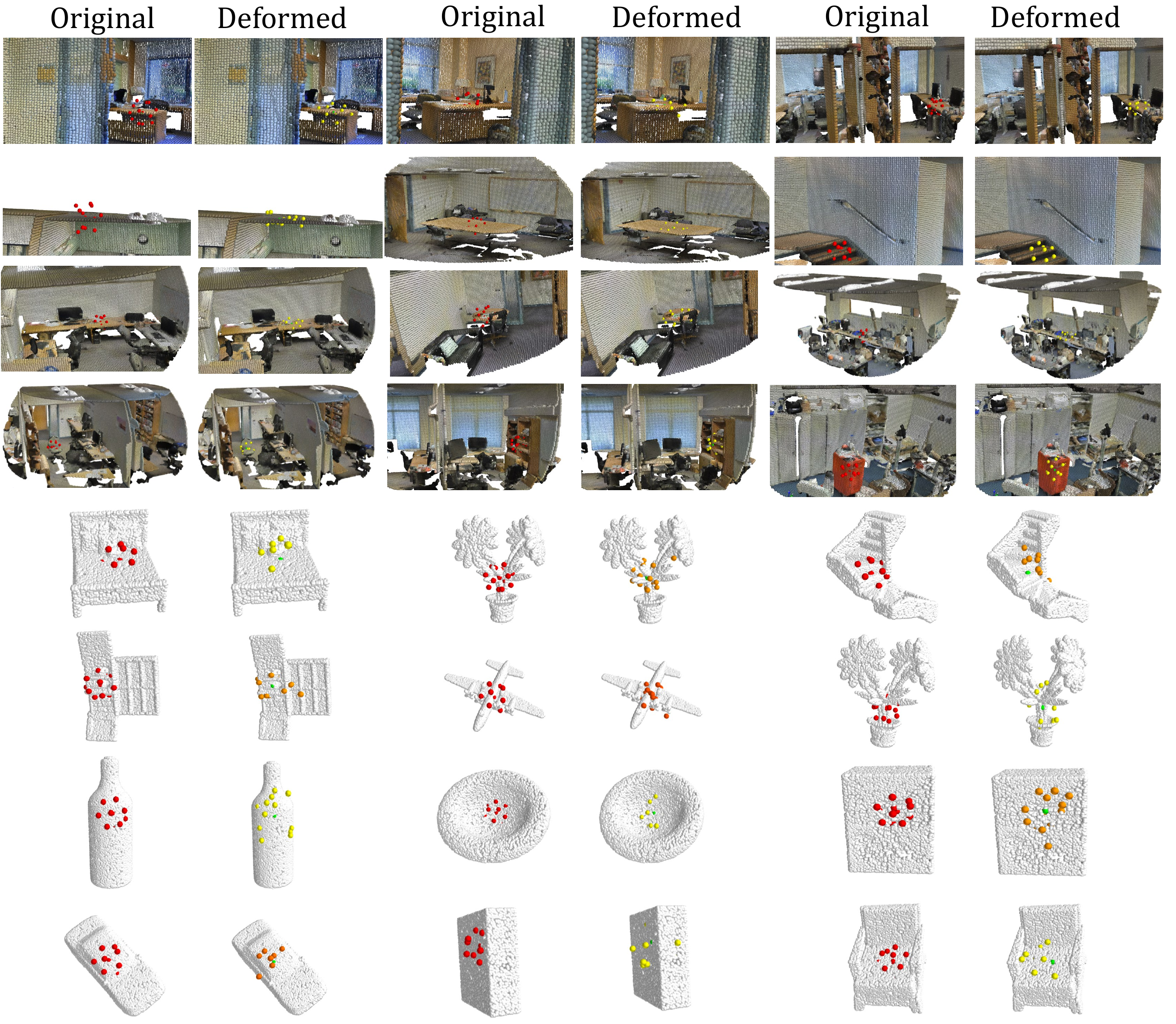}
\caption{Visualization of the deformable convolutional kernels in S3DIS (top four rows) and ModelNet (bottom four rows) respectively, please zoom in for details.}
\label{deform_s3dis}
\end{figure}
\begin{figure}[ht]
\centering
\includegraphics[scale=0.23]{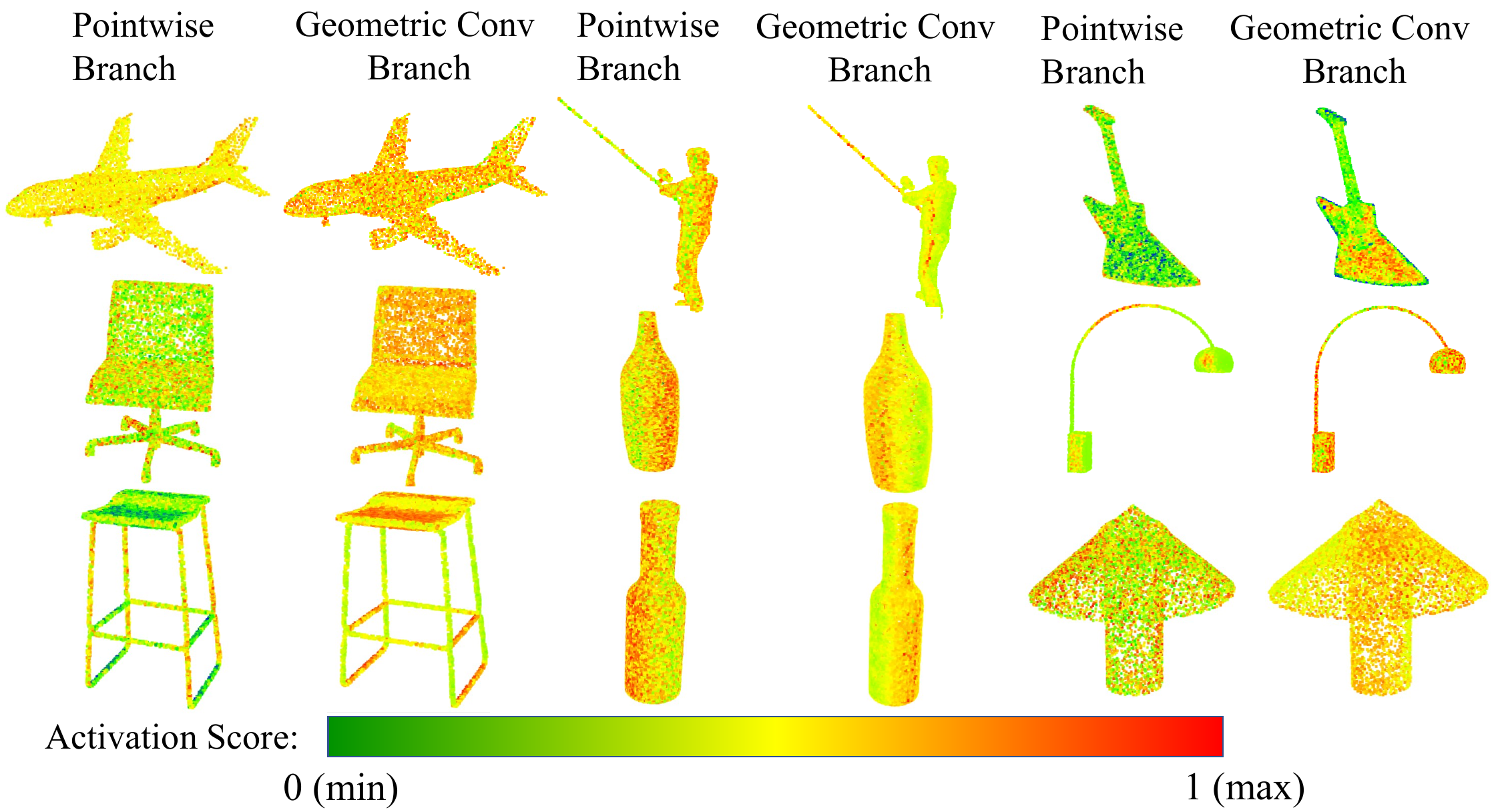}
\caption{Visualization of the learnt features by pointwise correlated feature learning and geometric convolutions respectively, please zoom in for details.}
\label{fig_activation}
\end{figure}
\begin{figure}[ht]
\centering
\includegraphics[scale=0.25]{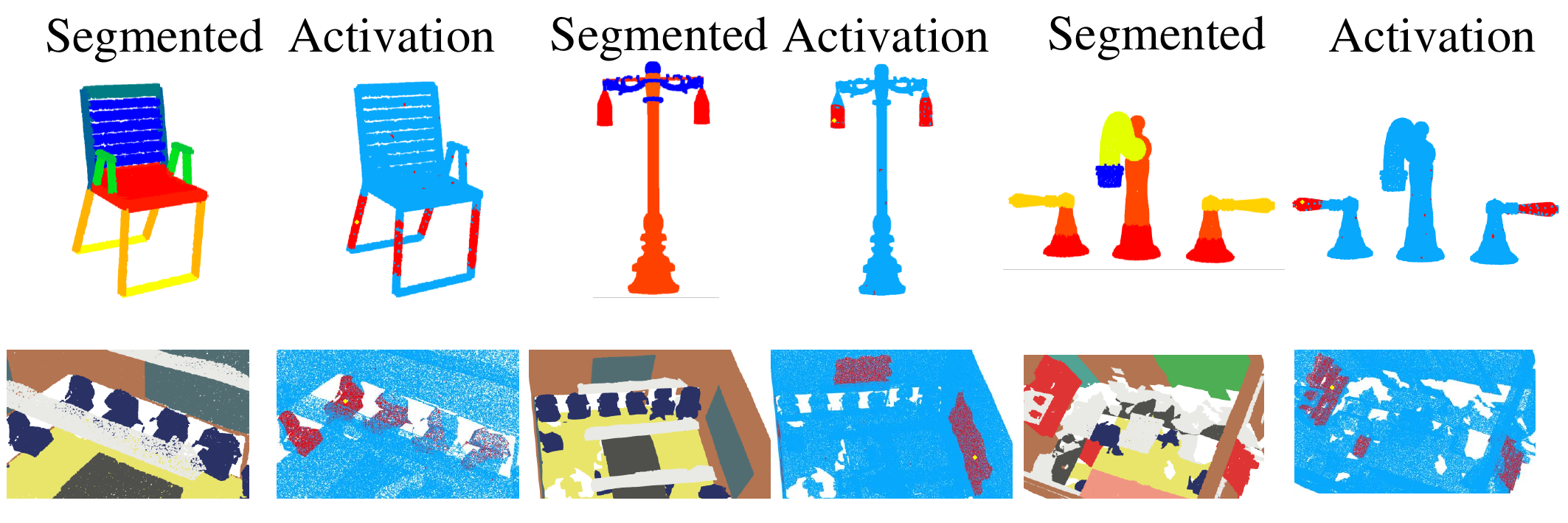}
\caption{Visualization of the non-local activations. Left shows the segmentation predictions; Right shows the non-local activations. (The background is indicated in blue and the query point is indicated in yellow, and the red points are given large attentive weights), please zoom in for details.}
\label{fig_acti}
\end{figure}
\begin{figure}[ht]
\centering
\includegraphics[scale=0.58]{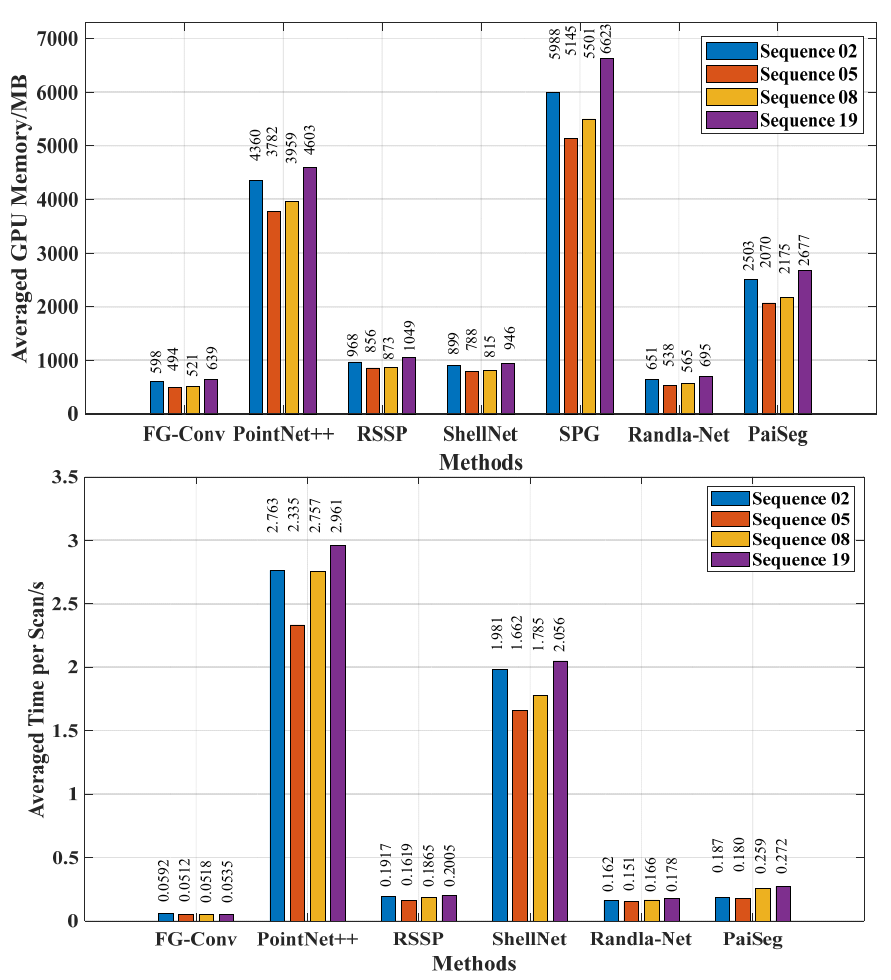}
\caption{Comparisons of the time and memory cost of different methods.}
\label{fig_eff}
\vspace{-5mm}
\end{figure}
To compare the efficiency our proposed \textit{\textbf{IGSAM}} with different sampling methods, we have experimented their GPU memory usage and processing time on a single GTX 1080 GPU with 8 GB memory. The sampling methods include Random Sampling (RS), Reinforcement Learning based Sampling (RLS) \cite{stadie2018importance}, GSS \cite{maddison2016concrete}, IDS, Farthest Point Sampling (FPS), and Generative Network (GS) \cite{lang2020samplenet} based Sampling. The point clouds are divided into batches consisting of $10^2, 10^3, 10^4, 10^5, 10^6$ and $10^7$ points respectively, then the batches of points are down-sampled 5 times which imitates the down-sampling in our network shown in Fig. \ref{fig_resnet}. The total time and memory consumption of sampling methods on different numbers of points are illustrated in Fig. \ref{fig_comp}. It can be demonstrated that RS has the fastest processing speed with the smallest memory consumption. However, RS will result in a stochastic loss in meaningful information, which will give unsatisfactory segmentation results. As shown in Table \ref{table_sampling}, mIOUs will drop significantly from $70.8\%$ to $66.8\%$ if RS is adopted. It should also be noted that GSS is not suitable for more than $10^6$ points because the GPU memory will increase greatly with the number of points. Hence, we only use GSS in the last layer of the network when the number of points is less than $10^5$. Leveraging IDS with adaptability to the local density of points, our \textit{\textbf{IGSAM}} achieve the best performance among different sampling methods with only a marginal increase of computational cost compared with RS.
The segmentation mIOUs using different sampling methods are also shown in Table \ref{table_sampling}. It can be seen that our sampling methods give the best performance among all sampling methods on the S3DIS benchmark with mIOUs of 70.8\%, which demonstrates the effectiveness of our proposed sampling strategy.

\subsection{Experiments of Large-scale Scene Understanding}
We have experimented our method extensively on nearly all existing large-scale point clouds understanding datasets including ModelNet40\cite{wu20153d}, ShapeNet-Part\cite{yi2016scalable}, PartNet\cite{mo2019partnet}, S3DIS\cite{armeni20163d}, NPM3D\cite{roynard2018paris}, Semantic3D\cite{hackel2017semantic3d}, Semantic-KITTI\cite{behley2019semantickitti}, and Scannet\cite{dai2017scannet}. The detailed information of different datasets and the results with speed of our framework in point clouds understanding are shown in Table \ref{tab_benchmark}. The qualitative experiments of large-scale real-world scene parsing are shown in Fig. \ref{fig_partnet}, Fig. \ref{fig_Sem3D}, Fig. \ref{fig_s3dis_detail}, Fig. \ref{fig_s3dis}, Fig. \ref{fig_Skitti}, and Fig. \ref{fig_NPM3D}, respectively. The mIOUs of 78.2\%, 70.8\%, 82.3\%, 58.2\% are attained on Semantic3D \cite{hackel2017semantic3d}, S3DIS \cite{armeni20163d}, NPM3D \cite{roynard2018paris} and fine-grained segmentation dataset PartNet \cite{mo2019partnet} respectively with real time segmentation of about 18.6 Hz for each LiDAR scan with $5 \times 10^5$ points, which outperforms state-of-the-art methods in terms of accuracy and efficiency. Denote the network with global attentive module in Fig. \ref{fig_nonlocal} as \textit{\textbf{FG-Net-v0}} and the one without as \textit{\textbf{FG-Net-v1}}. With global attention, the feature correlations across a long spatial range are effectively captured and distributed objects such as road and buildings are more precisely segmented in \textit{\textbf{FG-Net-v1}} compared with \textit{\textbf{FG-Net-v0}}. The transfer learning results shown in Fig. \ref{fig_transfer} also demonstrates our networks learn the underlying latent model of feature representations that is able to generalize across new scenes.
\begin{table*}[htbp!]
\caption{The comparisons of performance of our method on different large-scale point clouds understanding datasets}
\label{tab_benchmark}
\begin{center}
\begin{tabular}{cccc}
\toprule
Datasets & Main Features& Accuracy/ mIOUs (\%) & Time (s)/ $10^5$ points\\

\hline
ModelNet40\cite{wu20153d}& More than 12311 CAD models from 40 classes &93.1& 0.0561\\ 

ShapeNet-Part\cite{yi2016scalable}&16,881 shapes from 16 classes, annotated with 50 parts in total&86.6&0.0552\\


PartNet\cite{mo2019partnet}&More than 27000 3D models in 24 object classes with 18 parts per object &58.2&0.0501\\

S3DIS\cite{armeni20163d}&LiDAR scan of more than 6000 $m^2$ of 6 different areas from 3 buildings&70.8&0.0498\\

NPM3D\cite{roynard2018paris}&Kilometer-scale point clouds captured from multiple city roads&82.3&0.0528\\

Semantic3D \cite{behley2019semantickitti} & Largest-scale dataset of more than 4 billion annotated points with 8 classes&78.2&0.0523\\

Semantic-KITTI \cite{behley2019semantickitti}&Largest-scale LiDAR dataset for autonomous driving&53.8&0.0512\\

Scannet\cite{dai2017scannet}&Reconstructed 1513 indoor scenes from 707 different areas&68.5&0.0495\\

\bottomrule
\end{tabular}
\end{center}
\vspace{-3mm}
\end{table*}

\subsection{Visualization of the Network Modules}
\subsubsection{Visualization of the Deformable Convolutional Kernels}

To better demonstrate the geometry adaptive capacity of the deformable convolutions, the deformable kernel are visualized in  Fig. \ref{deform_s3dis}. It can be seen that kernel points are adaptively deformed to capture different geometric structures in the original point clouds. Hence in the test phase, the specific geometric structures in the unseen scene will be effectively captured and described by deformable kernels. In this way, we can model the geometry of the scene in a learnable way to better enhance structural awareness in per-point based processing.
\subsubsection{Visualization of the learnt features}

 To figure out what has been learnt by our network, the inner activations of the 2 core network modules are visualized as is illustrated in Fig. \ref{fig_activation}. It should be noted that the activations of designed 2 branches is complementary to each other because the deformable convolution captures the coarse-grained continuous geometry features while the correlated features learning focuses on fine-grained isolated per-point features as shown in Fig. \ref{fig_activation}, which is also in accordance with our design.
\subsubsection{Visualization of the non-local activation}

In order to further demonstrate the effectiveness of the non-local module, we also visualize the non-local activation. Fig. \ref{fig_acti} demonstrates that the non-local module can capture the long-range dependencies of the same semantic category such as chairs or bookcases. The contexts that are far away from each other can be nicely modeled and captured. It can also be observed that the non-local activation can also give rough results of segmentation prediction of the category of the query point, which is advantageous for further semantic segmentation tasks.
\vspace{-7mm}
\subsection{Efficiency and Online Performance}

To give a more convincing evaluation of the real-time performance of our method on large-scale real-scene point clouds segmentation, we have tested our method compared with others on the whole sequence of Semantic-KITTI \cite{behley2019semantickitti} dataset. The sequences are captured and fed into the networks at 25 Hz. The inference time and the GPU memory used are shown in Fig. \ref{fig_eff}, and the PaiSeg \cite{gao2020pai} is also a recently developed method for point clouds segmentation. With the residual learning framework, our method can reach 16.89 Hz, 19.53 Hz, 19.31 Hz, and 18.69 Hz for LIDAR scan 02, 04, 05, and 09 respectively. Compared with RSSP\cite{wang2019re} and RandlaNet \cite{hu2020randla}, the speed increase by 274\% and 38.5\% while the memory consumption reduce 46.5\% and 8.6\% respectively, which is a prominent progress in the speed and memory efficiency.

\subsection{Ablation Study of Network Modules}
\begin{table}[bp!]
\caption{The segmentation performance of ablated network on S3DIS}
\label{ablation}
\begin{center}
\begin{tabular}{cc}
\toprule

Ablations & mIOUs (\%)\\

\hline
Remove pointwise feature relation mining (PFM)&66.2\\ 

Remove geometric convolutional modelling (GCM)&59.8\\

Remove attentional aggregation (AG)&67.1\\

Remove global feature extraction&63.5\\

The full network framework&\textbf{70.8}\\

Replace the backbone with a 6-stage network&69.1\\

Replace the backbone with a 4-stage network&67.9\\

Without semantic context loss $\textit{\textbf{L}}_2(\textit{\textbf{W}})$&68.2\\

With 2 RLB2 in each convolutional block&69.7\\

Choose $M=1$ in RLB1 and RLB2 &70.6\\
\bottomrule
\end{tabular}
\end{center}
\vspace{-2mm}
\end{table}
Our designed network modules can be easily integrated or removed from existing point clouds processing architectures. Some ablation studies are also done to validate the effectiveness and necessity of our designed modules. As shown in Table. \ref{ablation}, 4 core modules are removed from our network respectively and the mIOUs of 6-fold cross-validation on S3DIS dataset is recorded. From the results, removing geometric convolutional modelling results in 11\% performance drop because learning the intrinsic geometric shape contexts of point clouds is vital for the recognition. On the other hand, removing global and local correlated feature mining results in 5.6\% and 7.3\% drop in mIOUs, which demonstrates both the local and long-range feature relationship capturing are also essential to the segmentation task. And not using the attentional aggregation will also decline the performance for not retaining some meaningful features. Furthermore, the experiments show the 5-stage ($h=5$) network is superior to 4-stage or 6-stage networks because shallow networks have a poor fitting ability while deeper networks will result in oversampling of point clouds, which will all deteriorate the performance. We also tested $M=1$ in the RLB, however the segmentation performance will not increase. Therefore $M=8$ is adopted for the sake of memory efficiency.

\section{Conclusion}
In this work, we have proposed a general solution \textit{\textbf{FG-Net}} to large-scale point clouds understanding with real-time speed and state-of-the-art performance. The filtering and sampling methods are specially designed for large-scale point clouds with high efficiency and they will boost the scene parsing performance. The network can effectively model the point clouds structures and find the feature correlations across a long spatial range. Leveraging feature pyramid based residual learning, hierarchical features at different resolutions can be fused in a memory efficient way. Experiments on challenging circumstances showed that our approach outperforms state-of-the-art methods in terms of both accuracy and efficiency.

\ifCLASSOPTIONcaptionsoff
  \newpage
\fi



\bibliographystyle{IEEEtran}
\bibliography{references}
%




%
\begin{IEEEbiography}[{\includegraphics[scale=0.73,keepaspectratio]{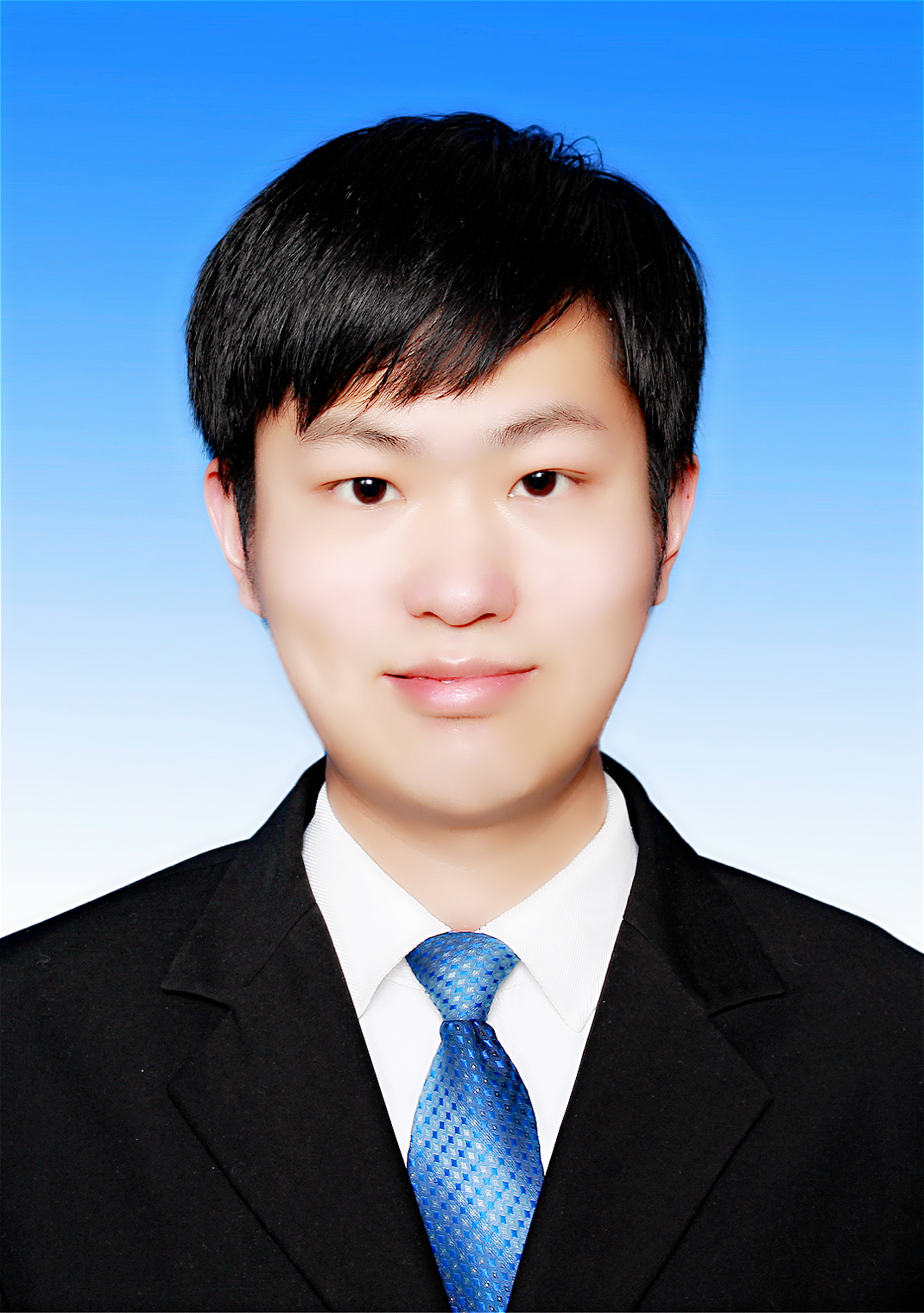}}]{Kangcheng Liu}
 received his B.Eng. degree in Electrical Engineering and Automation at Harbin Institute of Technology in 2018. He is currently pursing his Ph.D. degree in the Department of Mechanical and Automation Engineering, The Chinese University of Hong Kong. His research interests include power systems analysis and control, robotics, LIDAR-SLAM, machine learning and computer graphics for unmanned autonomous systems. Currently, he has strong interests in 3D deep learning and scene understanding for robotic perception.
\end{IEEEbiography}


\begin{IEEEbiography}[{\includegraphics[scale=0.225,keepaspectratio]{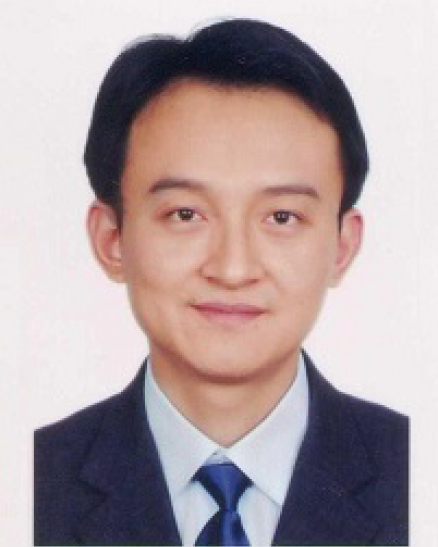}}]{Zhi Gao}
received the B.Eng. and the Ph.D degrees
from Wuhan University, China in 2002 and
2007 respectively. Since 2008, he joined the Interactive and Digital Media Institute, National University of Singapore (NUS), as a Research Fellow (A) and project manager. In 2014, he joined Temasek Laboratories in NUS (TL@NUS) as a Research Scientiest (A) and Principal Investigator. He is currently working as a full professor with the School of Remote Sensing
and Information Engineering, Wuhan University. He also serves as an associate editor of the journal Unmanned Systems.
Since 2019, he has been supported by the distinguished professor program of Hubei Province and the National Young Talent Program, China. He has published more than 70
research papers on top journals and conferences, such as IJCV, IEEE
T-PAMI, IEEE TIE, IEEE TGRS, IEEE T-ITS, ISPRS JPRS, Neurocomputing, IEEE TCSVT, CVPR, ECCV, ACCV, BMVC, etc. His research interests include computer vision, machine
learning, remote sensing and their applications. In particular, he has
strong interests in vision for intelligent systems and intelligent system based vision.
\end{IEEEbiography}

\begin{IEEEbiography}[{\includegraphics[scale=0.131,keepaspectratio]{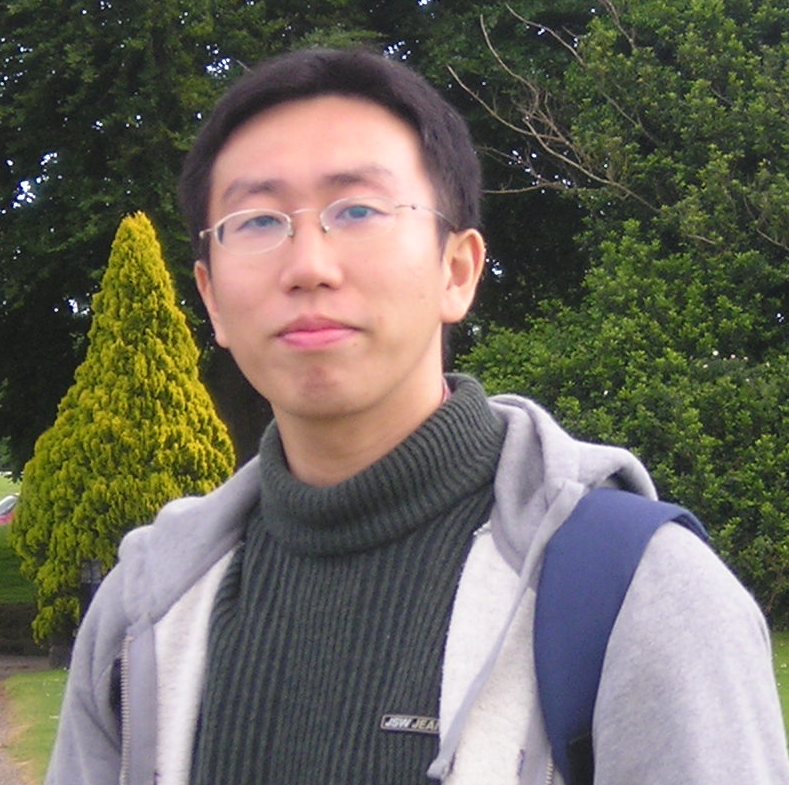}}]{Feng Lin}
 received the B.Eng. degree in Computer Science and Control, and the M.Eng. degree in system engineering from the Beihang University, Beijing, China, in 2000 and 2003, respectively. He received the Ph.D. degree in Computer \& Electrical Engineering from the National University of Singapore (NUS) in 2011. He was the recipient of the Best Application Paper Award, 8th World Congress on Intelligent Control and Automation, Jinan, China (2010).
Dr. Lin has worked as a Senior Research Scientist at the Temasek Laboratories \@ NUS, and a Research Assistant Professor of Department of Electrical \& Computer Engineering at the National University of Singapore from 2011 to 2019. 
Dr. Lin is currently working as an Associate Research Scientist at Peng Cheng Laboratory since 2019. His main research interests are unmanned aerial vehicles, vision-aided control and navigation, target tracking, robot vision as well as embedded vision systems. He has served on the editorial board for Unmanned Systems. 
\end{IEEEbiography}

\begin{IEEEbiography}[{\includegraphics[scale=0.1399, keepaspectratio]{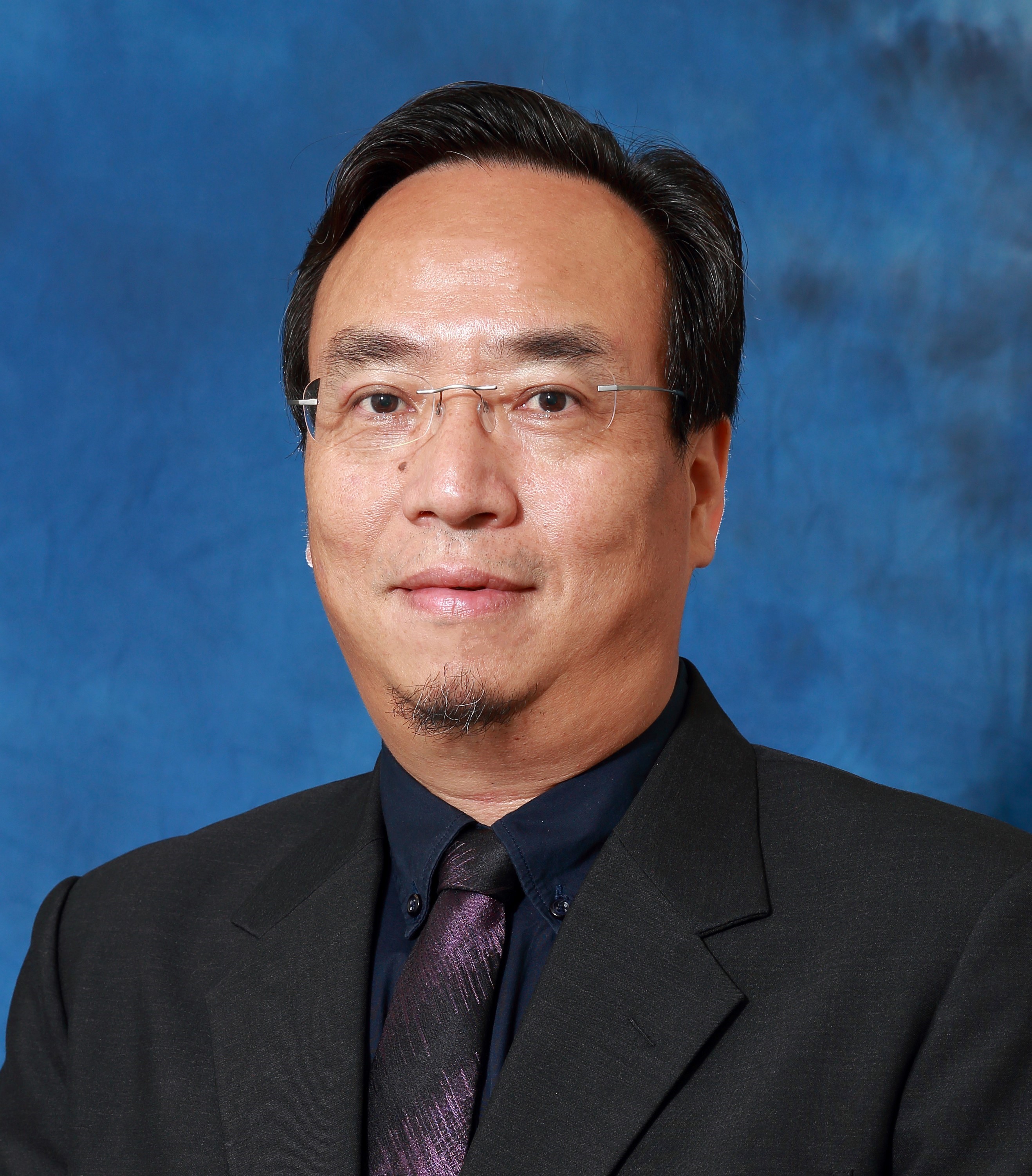}}]{Ben M. Chen} is currently a Professor of Mechanical and Automation Engineering at the Chinese University of Hong Kong (CUHK) and a Professor of Electrical and Computer Engineering (ECE) at the National University of Singapore (NUS). He was a Provost's Chair Professor in the NUS ECE Department, where he also served as the Director of Control, Intelligent Systems and Robotics Area, and Head of Control Science Group, NUS Temasek Laboratories. He was an Assistant Professor at the State University of New York at Stony Brook, in 1992–1993. His current research interests are in unmanned systems, robust control and control applications. 
Dr. Chen is an IEEE Fellow. He has authored/co-authored more than 450 journal and conference articles, and a dozen research monographs in control theory and applications, unmanned systems and financial market modeling. He had served on the editorial boards of a dozen international journals including IEEE Transactions on Automatic Control and Automatica. He currently serves as an Editor-in-Chief of Unmanned Systems. Dr. Chen has received a number of research awards. His research team has actively participated in international UAV competitions and won many championships in the contests.

\end{IEEEbiography}






\end{document}